%% file: main.tex
\documentclass[final]{siamltex}
\usepackage{algorithm,algpseudocode}
\usepackage{latexsym, graphicx, epsfig, amsmath, amsfonts,amssymb,bm}
\graphicspath{{Figures/}}
\usepackage{caption, subcaption}
\usepackage{epstopdf}
\usepackage{booktabs}
\usepackage{array}
\usepackage{color}
\usepackage{lineno}
\usepackage{url}

\newtheorem{remark}{Remark}[section]

\def\be{\begin{equation}}
\def\ee{\end{equation}}

\def\x{\mathbf{x}}
\def\ve{\mathbf{e}}
\def\y{\mathbf{y}}
\def\ty{\widetilde{\mathbf{y}}}
\def\f{\mathbf{f}}
\def\hf{\widehat{\mathbf{f}}}
\def\p{\mathbf{p}}
\def\q{\mathbf{q}}

\def\Y{\mathcal{Y}}
\def\X{\mathcal{X}}
\def\C{\mathcal{C}}

\def\E{{\mathcal E}}
\def\R{{\mathbb R}}

\newcommand{\argmax}{\operatornamewithlimits{argmax}}

\newcommand{\figref}[1]{Fig.~\ref{#1}}
\renewcommand{\eqref}[1]{(\ref{#1})}

\title{A Non-Intrusive Correction Algorithm for Classification Problems with Corrupted Data}

\author{Jun Hou, Tong Qin, Kailiang Wu \and Dongbin
       Xiu\thanks{Department of Mathematics,
		The Ohio State University, Columbus, OH 43210, USA.
		{\tt hou.345@osu.edu, qin.428@osu.edu, wu.3423@osu.edu, xiu.16@osu.edu.}
		Funding: This work was partially supported by AFOSR FA9550-18-1-0102.}
}

\begin{document}
\maketitle
\begin{abstract}
A novel correction algorithm is proposed for multi-class classification problems with corrupted 
training data. The algorithm is non-intrusive, in the sense that it post-processes a trained classification model by adding a correction procedure to the model prediction. The correction procedure can be coupled with any approximators, such as logistic regression, neural networks of various architectures, etc. When training dataset is sufficiently large, we prove that the corrected models deliver correct classification results as if there is no corruption in the training data.
For datasets of finite size, the corrected models produce significantly better recovery results, compared to the models without the correction algorithm.
All of the theoretical findings in the paper are verified by our numerical examples. 
\end{abstract}

\input Introduction

\input Setup

\input Analysis

\input Method

\input Extension

\input Examples

\input Conclusion
\input Appendix

\bibliographystyle{siamplain}
\bibliography{neural,labelCorrupt}

\end{document}

%% file: Introduction.tex
\section{Introduction} \label{sec:intro}

Classification problems arise in many practical applications, such as image classification, speech recognition, spam filtering, and so on. Over the past decades, classification has been widely studied by using machine learning techniques, which seek to learn a classifier from labeled training dataset to predict class labels for new data. 
However, real-world datasets often contain noise and their class labels can be corrupted, i.e., mislabelled. This can be caused by a variety of reasons, including human error, measurement error, or subjective bias by labelers, etc.
Label corruptions also occur in data poisoning \cite{li2016data,steinhardt2017certified}. For a more comprehensive review of the sources of label corruptions, see Section B of \cite{frenay2014classification}. 
Label corruptions, natural or malicious, can adversely impact classification performance of classifiers. See, for example, \cite{zhang2003robustness, zhu2004class, Nettleton2010} for impacts on
different machine learning techniques. 
It is therefore important to explore robust techniques that can mitigate, or even eliminate, the consequences of label corruptions.

\subsection{Related work}

There exist a large amount of literature on learning of classifiers in the presence of label noises/errors. See, for example, \cite{frenay2014classification} for 
a detailed survey. 
%
Methods to enhance model robustness against label noises include modifying network architecture and introducing corrections to loss function \cite{larsen1998design, patrini2017making, hendrycks2018using}. Larsen et al.~\cite{larsen1998design} proposed a framework for designing robust neural network classifiers by introducing a probabilistic model
for corruptions. Mnih and Hinton \cite{mnih2012learning} introduced two 
robust loss functions to deal with incomplete or poorly registered labels for binary classification of aerial images. 
In \cite{sukhbaatar2014training}, 
Sukhbaatar et al.~suggested introduction of a
noise layer into neural network models to adapt the network outputs to match the noisy label distribution. The
parameters of the noise layer was estimated as part of the training process and
involved modifications to current training infrastructures for deep network \cite{sukhbaatar2014training}. 
Later, Patrini et al.~\cite{patrini2017making} developed two procedures
for loss function correction, based on transition matrix measuring the probabilities of each class being corrupted into
another. 
They also proposed an estimate of those probabilities \cite{patrini2017making}, by extending the noise estimation technique in \cite{menon2015learning} 
to multi-class setting. 
The readers are also referred to \cite{xiao2015learning,veit2017learning,li2017learning,hendrycks2018using} for more studies on label noise robustness and loss correction techniques under the assumption that one has access to a small subset of clean data during training. 

Efforts were also made to design inherently noise-tolerating (also called noise-robust) algorithms or loss functions. 
For binary classification, it was proved that 0-1 loss is robust to 
symmetric or uniform label noises, while most of the standard convex loss functions are not
\cite{long2010random,manwani2013noise}. 
Several theoretically motivated noise-tolerating 
loss functions, including ramp loss, unhinged loss and savage loss, 
have been introduced in the context of support vector machines (cf.~\cite{brooks2011support,van2015learning,masnadi2009design}).
For binary classification, Natarajan et al.~\cite{natarajan2013learning} proposed an approach to modify any given surrogate loss function to achieve noise robustness.
In the context of deep neural networks, 
Ghosh et al.~\cite{ghosh2015making,ghosh2017robust} derived sufficient conditions for loss function to be robust 
against label corruptions for binary classification \cite{ghosh2015making} and multi-class classification \cite{ghosh2017robust}. 
Recently, Zhang and Sabuncu \cite{zhang2018generalized} 
generalized the commonly-used categorical 
cross entropy (CCE) loss to 
a set of noise-robust loss functions, which includes mean absolute error (MAE) loss as a special case.  
Other techniques that address various aspects of learning with noisy labels. 
They include, but are not limited to, cleaning up noisy labels \cite{veit2017learning,northcutt2017learning}, directly modelling the label noise and then using the expectation-maximization algorithm to learn the distribution of the true labels \cite{xiao2015learning,khetan2017learning}, and reweighting the samples according to the confidence in them \cite{ liu2016classification, ren2018learning, jiang2017mentornet}.

\subsection{Contributions of the present paper}

The focus of this paper is on a novel correction algorithm for multi-class classification problems with corrupted training data. 
A distinct feature of our algorithm is that the correction procedure is applied to the output of a pre-trained model. That is, it does not require modification to a particular model training method and  is performed only after the completion of the model training.
Therefore, our correction method is \textit{non-intrusive} and highly flexible for practical computations. The non-intrusive feature is not available for many of the aforementioned existing techniques
(cf.~\cite{larsen1998design,mnih2012learning,sukhbaatar2014training,patrini2017making,hendrycks2018using}), most of which require
modification to the model training architecture and/or loss function.
The proposed correction procedure in this paper, on the other hand, can be readily coupled with any existing classification methods, such as logistic regression or deep neural network learning, provided that categorical 
cross entropy (CCE) loss or
squared error (SE) loss is employed. 
The proposed correction algorithm is based upon our theoretical analysis for classification problems with corrupted dataset. 
We prove that, for sufficiently large dataset, the impact of corruption errors is minimal. More precisely, upon applying the proposed correction algorithm, the classification results
become exact, as if there is no data correction, when the size of dataset approaches infinity.
We also derive conditions, under which the original model without using the correction algorithm becomes inherently robust against label corruptions. Moreover, if  
the probability of mis-classification is uniform, we prove that the classification results are always correct in the limit of infinitely large dataset, provided that  a (small) portion of clean data exists in the dataset. 
Numerical examples are provided to confirm the theoretical analysis and 
demonstrate the performance of the proposed correction algorithm. 

This paper is organized as follows. After the basic problem setup in Section \ref{sec:setup}, we present some theoretical analysis on classification problem with corrupted labels in
Section \ref{sec:analysis}. Based on the analysis,
our non-intrusive correction algorithm is then presented in Section \ref{sec:method}. 
Extensions of the analysis and algorithm to more general cases are presented in Section \ref{eq:discussion}.
In Section \ref{sec:examples}, we present an extensive set of numerical examples, including 
well known benchmark problems using real-word datasets, to 
verify the theoretical findings and 
demonstrate the effectiveness of the proposed correction 
algorithms.

%% file: Setup.tex
\section{Problem Setup} \label{sec:setup}
Let $D_1, D_2, \ldots, D_n$ be $n$ non-overlapping regions in $\R^d$ with $D_i\cap D_j=\emptyset$ for $i\neq j$.
A feature set is defined to be $\X=\bigcup_{k=1}^n D_k$ and  equipped with a probability measure $\omega$.
%
Each feature $\x\in \X$ is associated with a label $\y(\x)$. We use one-hot encoding for the label, i.e., $\y(\x)=\ve_k$ if $\x \in D_k$, where $\ve_{k}\in \R^n$ is $n$-vector with value $1$ in its $k$th component and $0$ otherwise. Let $\Y=\{\ve_k\}_{k=1}^n$ denote the label set. 

We are given a sample set $S=\{\x_i\}_{i=1}^M$, which are i.i.d. drawn from $(\X, \omega)$. For each sample $\x_i$, let $\ty({\x_i})$ be its observed label, which may be corrupted and different from the true label $\y ({\x_i})$.
We assume a subset of the labels $S_c\subset S$ are corrupted and denote its proportion to the entire dataset as $\lambda\in [0, 1)$, i.e., $|S_c|=\lambda M$.
For each sample $\x_i\in S_c$, we assume its observed label $\ty({\x_i})$ is a realization of a random variable $\mathbf{Y}$ with distribution 
\begin{equation}\label{Yprob}
{\rm Prob}(\mathbf{Y}=\ve_j)=\alpha_j, \qquad j=1,\dots,n, 
\end{equation}
where $\alpha_j\in [0,1]$ and $\sum_{j=1}^n\alpha_j=1$. We assume that the corruption ratio $\lambda$ and the distribution $\{\alpha_j\}_{j=1}^n$ are available (or can be reliably estimated). However, no prior information is available about the corrupted subset $S_c$. 

Let $\C :=\{\p\in \R^n: p_i\geq 0,\, \sum_{i=1}^n p_i=1\}$ be the probability simplex, where $p_i$ is the probability of a feature to be in $D_i$. We seek to learn  
 a probability function $\f =(f_1,\dots,f_n): \X\rightarrow \C$ and define a classifier
\begin{equation}\label{DefClassifier}
\hf(\x)= \ve_i, \quad  i=\argmax_{j\in \{ 1,2,\ldots,n \} } f_j.
\end{equation}
If maximum probability is attained by multiple labels, we define the first one as the predicted classifier. 
The classification \textit{completely recovered}, if for any $\x\in \X$,  we have $\hf(\x)=\y({\x})$. That is, the classification is able to correctly produce the true classification.

We employ neural networks to train the classifier via minimizing the following empirical risk
\begin{equation}\label{risk}
	\E(\f):=\frac{1}{M}\sum_{i=1}^M L(\ty({\x_i}), \f(\x_i;\Theta)),
\end{equation}
where 
$\Theta$ denotes the model parameters in the network and $L: \Y\times {\mathcal C}\rightarrow \R^+ \cup \{ 0 \}$ is the loss function. We are interested in the commonly used categorical cross-entroy (CCE) and squared-error (SE) loss functions, defined as
\be\label{eq:cce_mse_loss}
L(\p, \q):=
\begin{cases}
	-\sum_{i=1}^n p_i \log q_i, & \text{CCE},\\
	\sum_{i=1}^n |p_i-q_i|^2, & \text{SE}.
\end{cases}
\ee
Let $\Theta^*$ be the network parameters upon satisfactory training and $\f (\x; \Theta^*)$ be the trained model. 


%% file: Analysis.tex
\section{Main Theoretical Analysis} \label{sec:analysis}


In this subsection, we present theoretical analysis for the above
classification problem. We first derive conditions on the corruption
ratio $\lambda$ and the distribution $\{\alpha_k\}_{1\leq k\leq n}$,
under which the CCE and the SE loss functions are inherently robust
against the label corruption.  Based upon the analysis, we propose a
modified classifier, to be used after data training, to
eliminate the impact of corrupted data. 

\subsection{Asymptotic Empirical Risk}
Most of analysis is based on the assumption that the data set $S$ is sufficiently large, i.e., $M\gg 1$. 
Let $S_j := (S\setminus S_c) \cap D_j$, $j=1,\dots, n$. 
The empirical risk \eqref{risk} can be split into $n+1$ parts as 
\begin{equation*}
\begin{aligned}
{\mathcal E}( \f ) &= \sum_{j=1}^n \bigg(  \frac{1}{M}  \sum_{ \x_i \in S_j  } L ( \ty ( \x_i), \f ( \x_i) )  \bigg)
+ \frac{1}{M} 
\sum_{ \x_i \in S_c  } L ( \ty ( \x_i), \f ( \x_i) )
\\
&= \sum_{j=1}^n \bigg(  \frac{1}{M}  \sum_{ \x_i \in S_j  }  L ( {\bf e}_j, \f ( \x_i) )  \bigg)
+\frac{1}{M} 
\sum_{ \x_i \in S_c  } L ( \ty ( \x_i), \f ( \x_i) )
\\  
& =: \sum_{j=1}^n {\mathcal E}_j  + {\mathcal E}_c.
\end{aligned}
\end{equation*}

When$M \gg 1$, the following approximation holds: 
\begin{align}\label{app1}
\begin{aligned}
{\mathcal E}_j 
& \approx   (1-\lambda) \int_{D_j}   L ( {\bf e}_j, \f ( \x) )  d \omega_j, \quad 1\le j \le n,
\end{aligned}
\end{align}
where $d \omega_j =  \frac1{ \int_{D_j} d \omega } d \omega $.  
Note that for all $ \x_i \in S_c$, the label $ \ty ( \x_i)$ 
 is a realization of the random variable $\mathbf{Y}$ with the
 distribution \eqref{Yprob}. When $M \gg 1$, the summation in $
 {\mathcal E}_c$ can be considered as approximation to  expectation,
\begin{equation}\label{app2}
\begin{aligned}
{\mathcal E}_c \approx \lambda \sum_{j=1}^n  \int_{D_j} \bigg(  \sum_{k=1}^n \alpha_k 
L ( {\bf e}_k, \f ( \x ) )
\bigg) d \omega_j.
\end{aligned}
\end{equation}
Therefore, when $M \gg 1$, we have 
\begin{equation}
  \label{Jf}
\begin{split} 
{\mathcal E} (\f ) & \approx 
(1-\lambda) \sum_{j=1}^n \int_{D_j}  L ( {\bf e}_j, \f ( \x) )  d \omega_j
+ \lambda \sum_{j=1}^n  \int_{D_j} \bigg(  \sum_{k=1}^n \alpha_k 
L ( {\bf e}_k, \f ( \x) )
\bigg) d \omega_j
\\
& =  \sum_{j=1}^n \int_{D_j} \bigg( (1-\lambda) L ( {\bf e}_j, \f ( \x) )
+  \lambda  \sum_{k=1}^n \alpha_k 
L ( {\bf e}_k, \f (\x) )
\bigg ) d \omega_j
\\
& =: J( \f ).
\end{split}
\end{equation}

As the number of data $M \rightarrow +\infty$, the empirical risk ${\mathcal E} (\f )$ approaches $J( \f )$. Subsequently, we call $J( \f )$ {\it asymptotic empirical risk}. 

\subsection{Main Results}
Our main theoretical results are summarized as follows.

\begin{theorem} \label{thm:nd}
	For both the CCE and SE loss functions \eqref{eq:cce_mse_loss}, 
	the function $\f^* $ that minimizes the asymptotic empirical
        risk $J(\f)$ \eqref{Jf} is 
	\begin{equation}\label{minf:n}
	\f^* ( \x ) = \begin{cases}
	\left( 1-\lambda + \lambda \alpha_1 , 
	\lambda \alpha_2,
	\dots, 
	\lambda \alpha_n
	\right), \quad &{\rm if}~\x\in D_1 ,
	\\
	\cdots &
	\\
	\left( \lambda \alpha_1 , \dots
\lambda \alpha_{j-1}, 1-\lambda + \lambda \alpha_j, \lambda \alpha_{j+1}
\dots, 
\lambda \alpha_n
\right), \quad &{\rm if}~\x\in D_j ,	1<j<n,
	\\
	\cdots &
	\\
	\left(  \lambda \alpha_1 , 
	\lambda \alpha_2,
	\dots, 
	1-\lambda + \lambda \alpha_n 
	\right), \quad &{\rm if}~\x\in D_n .		
	\end{cases}
	\end{equation}	
\end{theorem}

The proof of Theorem \ref{thm:nd} can be found in Appendix \ref{app1:proof}.

\begin{remark}
Theorem \ref{thm:nd} implies that the classification is completely recovered if and only if 
\begin{equation}
1-\lambda + \lambda \alpha_j > \lambda \max_{k \ne j} \{\alpha_k\}, \qquad j=1,\dots,n,
\end{equation}
or equivalently, 

\begin{equation}\label{ReCondition}
	\lambda < \frac{1}{ 1+ \max_{k} \{\alpha_k\}-\min_{k}\{\alpha_k\}}.
\end{equation}
In other words, under the condition \eqref{ReCondition}, it holds $\hf^* (\x) = \y( \x)$ for all $\x \in {\mathcal X}$.
\end{remark}

\begin{remark}
	Since $0\leq \alpha_k\leq 1$ for any $k$, a direct consequence of the condition \eqref{ReCondition} is that when $\lambda<\frac 12$, the classification can always be completely recoverred, given any corruption distribution $\{\alpha_k\}$.
\end{remark}

As a direct consequence of the above theorem, we have the following results for two special cases. 

\begin{corollary}\label{cor:2d}
Consider binary classification problem with  $n=2$ and corruption
probability $\alpha_1=\alpha$ and $\alpha_2=1-\alpha$.  For both the
CCE and SE loss functions \eqref{eq:cce_mse_loss}, 
the function $\f^* $ that minimizes the asymptotic empirical risk
$J(\f)$  \eqref{Jf} is 
			\begin{equation}\label{minf:2}
	\f^* ( x ) = \begin{cases}
	\left( 1-\lambda + \lambda \alpha, \lambda (1-\alpha) \right), \quad & \x \in D_1,
	\\
	\left( \lambda \alpha,  1- \lambda \alpha \right), \quad & \x \in  D_2.
	\end{cases}
	\end{equation}
The classification is completely recovered if and only if 
\begin{equation}\label{ReCon2}
1-\frac{ 1}{2 \lambda} < \alpha < \frac{1 }{2 \lambda},
\end{equation} 
which means $\hf^* (\x) = \y( \x)$ for all $\x \in {\mathcal X}$.
\end{corollary}

\begin{corollary}\label{cor:eqA}
Consider classification problem with symmetric corruption probability, i.e.,
	\begin{equation}\label{eqProb}
	\alpha_1=\alpha_2 = \dots= \alpha_n=\frac{1}{n}, \qquad n\geq 2.
	\end{equation}
Then, for both the CCE and SE loss functions \eqref{eq:cce_mse_loss},  
the function $\f^* $ that minimizes the asymptotic empirical risk
$J(\f)$ \eqref{Jf} is
		\begin{equation}\label{minf:nsp}
\f^* ( \x ) = \begin{cases}
\left( 1-\lambda + \frac{\lambda}n , 
\frac{\lambda}n ,
\cdots, 
\frac{\lambda}n
\right), \quad & {\rm if}~ \x \in D_1,
\\
\cdots&
\\
\left(  \frac{\lambda}n, 
\cdots, 
\frac{\lambda}n,
1-\lambda + \frac{\lambda}n
\right), \quad & {\rm if}~ \x \in D_n.
\end{cases}
\end{equation}
If $0\le \lambda < 1$, then $ \hf^* ( \x) = \y (\x)$ for all $ \x \in \mathcal X$, and the classification is completely recovered. 
\end{corollary}

\subsection{Post-Modified Classifier}
The analysis in the previous section suggests a way to modify the
trained classifier, so that the classification can become completely
recovered, even if label corruptions do not satisfy the condition \eqref{ReCondition}.
We refer this as post-modified classifier, because it can
be applied after the training is completed.

When the corruption probability $\{\alpha_k\}_{1\le k \le n}$ and
proportion $\lambda$ are known, via certain estimation procedure such
as \cite{sukhbaatar2014training,patrini2017making,hendrycks2018using},
we propose to use the following modified classification function:
\begin{equation}\label{fmod}
\f^{mod} (\x) =  \f^* ( \x) + \f_\Delta, 
\end{equation}
where $\f^* $ minimizes the asymptotic empirical risk
$J(\f)$ \eqref{Jf}  and $\f_\Delta$ is constant vector
\begin{equation}
\label{eq:correction}
\f_\Delta := \left( { \lambda \Big(\frac1n - \alpha_1 \Big) } , 
{ \lambda \Big(\frac1n -\alpha_2 \Big) },
\dots, 
{ \lambda \Big(\frac1n -\alpha_n \Big) }
\right).
\end{equation}

As a direct consequence of Theorem \ref{thm:nd}, we have the following conclusion.

\begin{theorem}\label{thm:mod}
	For both the CCE and SE loss functions \eqref{eq:cce_mse_loss}, the modified classification function defined in \eqref{fmod} satisfies
			\begin{equation}\label{fmod:n}
	\f^{mod} ( \x ) = \begin{cases}
	\left( 1-\lambda + \frac{\lambda}n , 
	\frac{\lambda}n ,
	\cdots, 
	\frac{\lambda}n
	\right), \quad & {\rm if}~ \x \in D_1,
	\\
	\cdots &
	\\
	\left(  \frac{\lambda}n, 
	\cdots, 
	\frac{\lambda}n,
	1-\lambda + \frac{\lambda}n
	\right), \quad & {\rm if}~ \x \in D_n.
	\end{cases}
	\end{equation}
	If $0 \le \lambda < 1$, the classifier $\hf^{mod} ( \x)$ associated with the function $\f^{mod} ( \x )$ satisfies  
	$$ \hf^{mod} ( \x) = \y (\x), \qquad \x \in \mathcal X.$$ 
\end{theorem}

Theorem \ref{thm:mod} indicates that 
the modified classifier $\hf^{mod} $ can completely recover the exact classification for any $\{\alpha_k\}_{1\le k \le n}$ and any $0\le \lambda< 1$.

\begin{remark}
For symmetric corruption probability
$\alpha_1=\alpha_2=\dots=\alpha_n=\frac1n$, we have $\f_\Delta = {\bf
  0}$.  The modified classification function \eqref{fmod} becomes the
unmodified one.  This is consistent with the result of Corollary \ref{cor:eqA}. 
\end{remark}

%% file: Method.tex
\section{Implementation and Extension}

\subsection{Implementation Algorithm}\label{sec:method}

In this section, we discuss implementation detail of the aforementioned classification method. Note that the theoretical analysis in the previous section does not depend on the type of
approximation for $\f$ -- it can be linear regression, nonlinear neural networks, etc. Our discussion here is in the context of neural network (NN), because it is the predominant methods used for classification problems, see, for example, \cite{graves2013speech, he2016deep}.

\begin{figure}[htbp]
	\centering
	\includegraphics[width=0.99\textwidth]{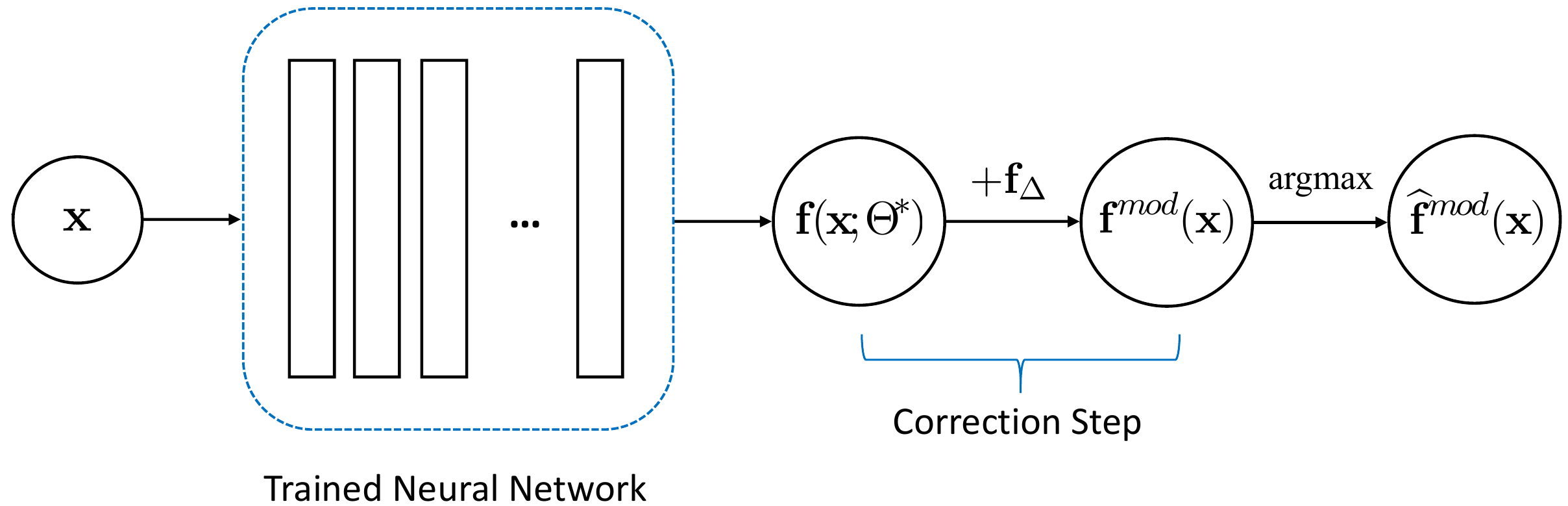}	
	\caption{\small
		Algorithm flowchart.
	}\label{fig:algorithm}
\end{figure}

Assume we are given a sample set $S=\{{\bf x}_i\}_{i=1}^M$, the corresponding observed labels $\{ \ty ({\bf x}_i) \}_{i=1}^M$, 
the corruption ratio $\lambda$ and the corruption distribution $\{ \alpha_j \}_{j=1}^n$. 
As illustrated in Fig.~\ref{fig:algorithm}, the implementation of our algorithm is outlined as follows. 

\begin{description}
	\item[Step 1] Construct a neural network (NN) with $\x$ as  input to approximate the probability function. Let $ {\bf f}: \mathbb R^d \to \mathbb R^n $ be the operator of the NN, where $\Theta$ is the parameter set including all the parameters in the network. 
	 To ensure the output always belongs to the probability simplex $\C$, we exploit the standard softmax function $\sigma ( {\bf z} )$, defined by 
	 $$
	 \sigma ({\bf z})_j =  \frac{ e^{z_j} }{ \sum_{i=1}^n e^{z_i} }, 
	 $$
	  in the output layer of the network. 
	\item[Step 2] Train the network via minimizing the empirical risk \eqref{risk} with either CCE or SE loss function  \eqref{eq:cce_mse_loss}. Let $\Theta^*$ be the trained parameters and $ {\bf f} (\x \Theta^*) $ be the trained network.
	\item[Step 3] Employ the non-intrusive post-correction \eqref{fmod}
          \begin{equation*}
	 {\bf f}^{mod} (\x) = {\bf f} (\x; \Theta^*) +  \f_\Delta.
       \end{equation*}
	\item[Step 4] Build the final classifier by argmax procedure:
\begin{equation}\label{DefC}
\hf^{mod} (\x)= \ve_i, \quad  i=\argmax_{j\in \{ 1,2,\ldots,n \} }  f^{mod}_j.
\end{equation}
\end{description}

A graph illustrating the steps is in Figure \ref{fig:algorithm}.

 Note that Step 3 and Step 4 are applied only after the network training has been completed. Therefore, they are ``non-intrusive'' and do not require modification to the NN structure or training. These steps can be applied to any suitable NN for classification problems.

%

%% file: Extension.tex
\subsection{Extension} \label{eq:discussion}

Our main theoretical results from Section \ref{sec:analysis} can be
extended to a more general setting, by extending the basic
assumption of the corruption probability \eqref{Yprob} 
More specifically,  we assume that, for each sample $\x_i\in S_c$, its observed label $\ty({\x_i})$ is a realization of a random variable $\mathbf{Y}$ with the distribution 
\begin{equation}\label{Yprob1}
{\rm Prob}(\mathbf{Y}=\ve_j ~|~ \y({\x_i}) = \ve_i )=\alpha_{i,j}, \qquad 1\le i,j \le n.
\end{equation}
This is the probability that the label ${\bf e}_i$ is corrupted to
${\bf e}_j$ and satisfies the obvious condition
$$
\sum_{j=1}^n \alpha_{i,j} = 1, \qquad 1\le i \le n.
$$
Let ${\bf A} :=(\alpha_{i,j})$ be the corruption probability matrix. and we have the following result, as
an extension of Theorem \ref{thm:nd}.
\begin{theorem} \label{thm:nd23}
	For both the CCE and SE loss functions
        \eqref{eq:cce_mse_loss}, the function $\f^* $ that minimizes
        the asymptotic empirical risk $J(\f)$  \eqref{Jf} is
	\begin{equation*}
\begin{aligned}
	\f^* ( \x )  &= 
	\begin{cases}
	\left( 1-\lambda + \lambda \alpha_{1,1} , 
	\lambda \alpha_{1,2},
	\dots, 
	\lambda \alpha_{1,n}
	\right), \quad &{\rm if}~\x\in D_1 ,
	\\
	\cdots &
	\\
	\left( \lambda \alpha_{j,1} , \dots
	\lambda \alpha_{j,j-1}, 1-\lambda + \lambda \alpha_{j,j}, \lambda \alpha_{j,j+1}
	\dots, 
	\lambda \alpha_{j,n}
	\right), \quad &{\rm if}~\x\in D_j ,	1<j<n,
	\\
	\cdots &
	\\
	\left(  \lambda \alpha_{n,1} , 
	\lambda \alpha_{n,2},
	\dots, 
	1-\lambda + \lambda \alpha_{n,n} 
	\right), \quad &{\rm if}~\x\in D_n .		
	\end{cases}
	\\[2mm]
	&=
	\left( (1-\lambda) {\bf I} + \lambda {\bf A}^\top \right)  \y (\x), \qquad \forall \x \in \mathcal X.
\end{aligned}
	\end{equation*}
\end{theorem}

\begin{proof}
	The proof is similar to the proof of Theorem \ref{thm:nd} and is omitted.
\end{proof}

Suppose the corruption probability matrix $\bf A$ and corruption
proportion $\lambda$ are known, via statistical estimation procedures
such as those in
\cite{sukhbaatar2014training,patrini2017making,hendrycks2018using},  and matrix $(1-\lambda) {\bf I} + \lambda {\bf A}^\top$ is nonsingular, we propose the following new modified classification function
\begin{equation*}
\f^{mod} (\x) = \left( (1-\lambda) {\bf I} + \lambda {\bf A}^\top \right)^{-1} \f^* (\x).  
\end{equation*}
This modified classification function can completely recover the exact classification for any $\lambda \in [0,1)$ and any transition matrix ${\bf A}$ satisfying $\det( (1-\lambda) {\bf I} + \lambda {\bf A}^\top) \neq 0$.

%% file: Examples.tex
\section{Numerical Examples} \label{sec:examples}
In this section, we present numerical examples. We first present two
well studied  academic examples to verify the theoretical results in
Section \ref{sec:analysis}. We then present two practical examples
using well known existing datasets to demonstrate the applicability of
the proposed algorithm on practical classification problems. In all
examples, we test both the CCE and the SE loss functions
\eqref{eq:cce_mse_loss}. Neural network training and cost minimization
problems are solved with the Adam algorithm \cite{kingma2014adam} with
the parameters set as in the Algorithm 1 of \cite{kingma2014adam}. All
the examples are implemented with the open-source libraries Keras
\cite{chollet2015keras} and Tensorflow
\cite{tensorflow2015-whitepaper}. 

\subsection*{Example 1: Binary Classification}
\begin{figure}[!htb]
	\centering
	\includegraphics[scale=0.6]{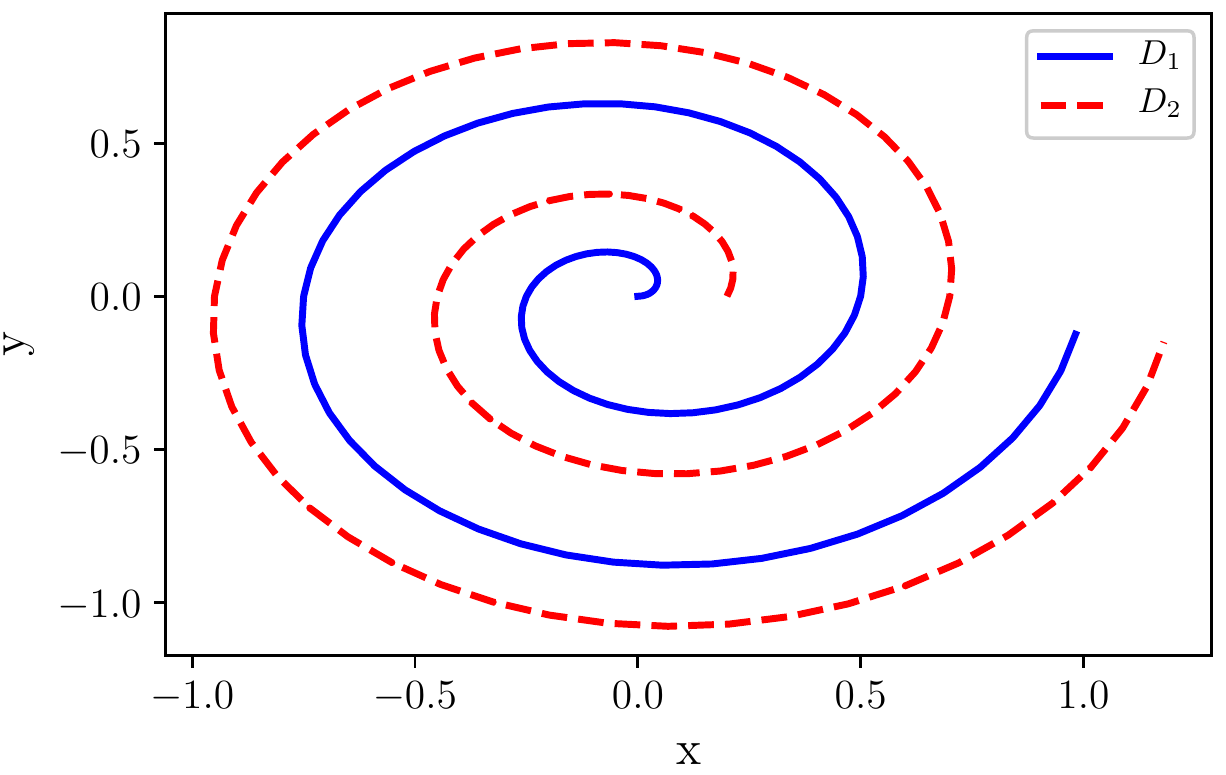}
	\caption{The illustration of the swiss roll example.}
	\label{fig:swiss_roll}
\end{figure}

We first consider a binary classification problem from the Swiss Roll
example \cite{haber2017stable}. The feature set $D=D_1\bigcup D_2$
consists of two spirals, as in \figref{fig:swiss_roll}, in the
following form
\begin{equation}
\label{eq:swiss_roll_equation}
D_1: 
\begin{cases}
x=r\cos(4\pi r),\\
y=r\sin(4\pi r),
\end{cases}
\quad\quad
D_2:
\begin{cases}
x=(r+0.2)\cos(4\pi r),\\
y=(r+0.2)\sin(4\pi r),
\end{cases}
\end{equation}
where $r\in [0, 1]$. The training samples are obtained by sampling the
parameter $r$ according to the uniform distribution over $[0, 1]$. To
verify the theories in in Section \ref{sec:analysis}  for sufficiently
large datasets, we use two million samples, with one million for each
class. The labels are then corrupted with a fixed corruption ratio $\lambda=0.7$. 

The classifier is constructed by employing a feedforward neural network with $3$ hidden layers and each layer contains $20$ neurons. We use the rectified linear unit (ReLU) \cite{nair2010rectified} activation function in the hidden layer and the softmax activation function in the output layer.

\begin{figure}
	\centering
	\begin{subfigure}[b]{0.49\textwidth}
		\centering
		\includegraphics[width=0.9\linewidth]{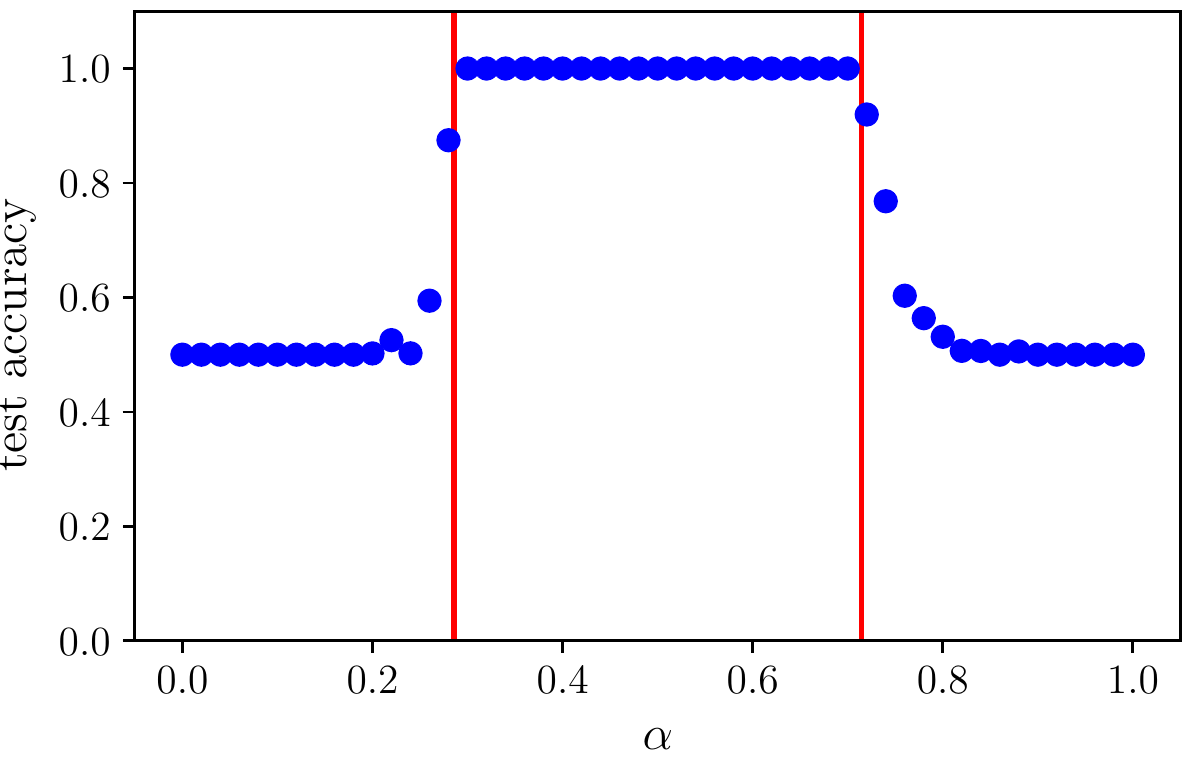}
		\caption{CCE loss, without the correction}
	\end{subfigure}
	\begin{subfigure}[b]{0.49\textwidth}
		\centering
		\includegraphics[width=0.9\linewidth]{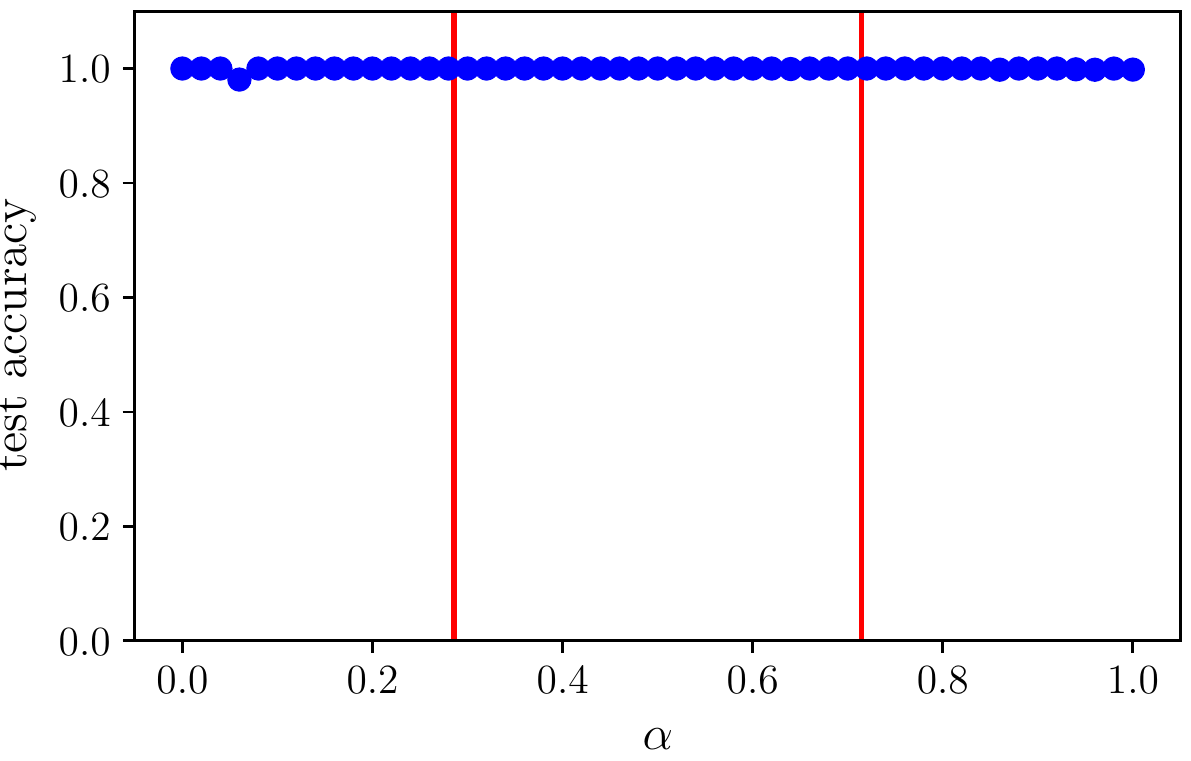}
		\caption{CCE loss, with the correction}
	\end{subfigure}
	\begin{subfigure}[b]{0.49\textwidth}
		\centering
		\includegraphics[width=0.9\linewidth]{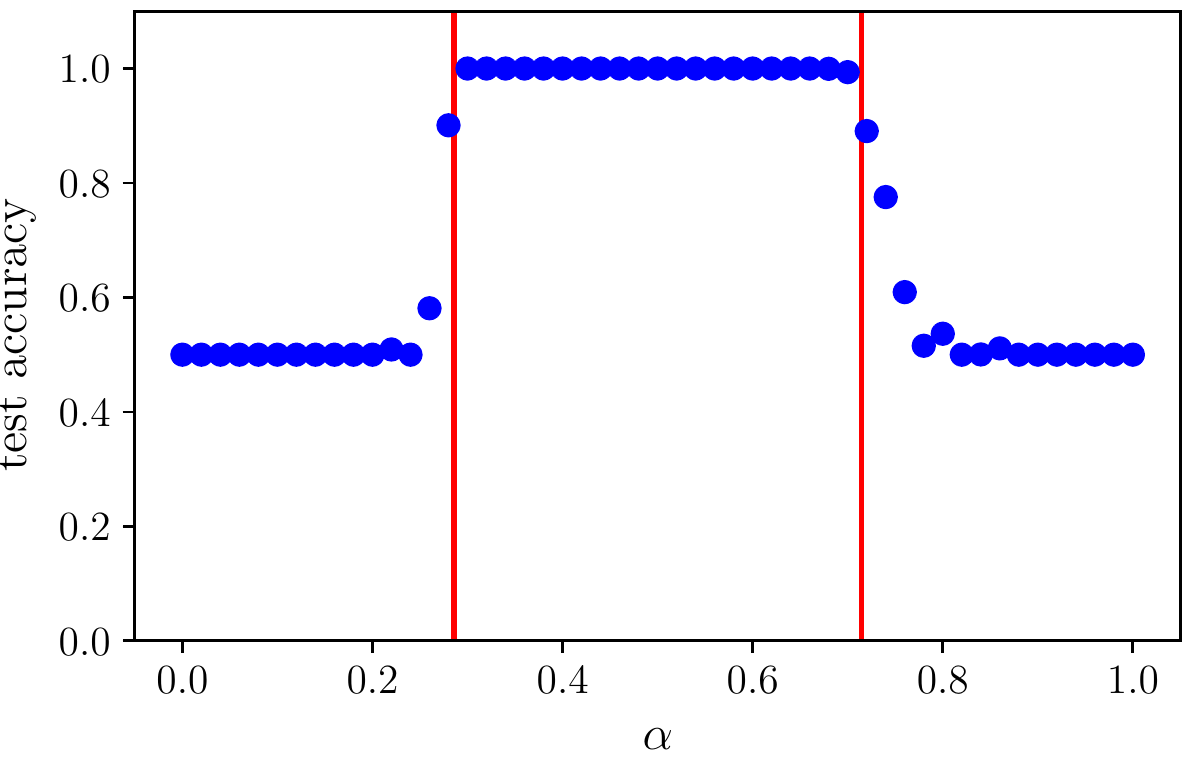}
		\caption{SE loss, without the correction}
	\end{subfigure}
	\begin{subfigure}[b]{0.49\textwidth}
		\centering
		\includegraphics[width=0.9\linewidth]{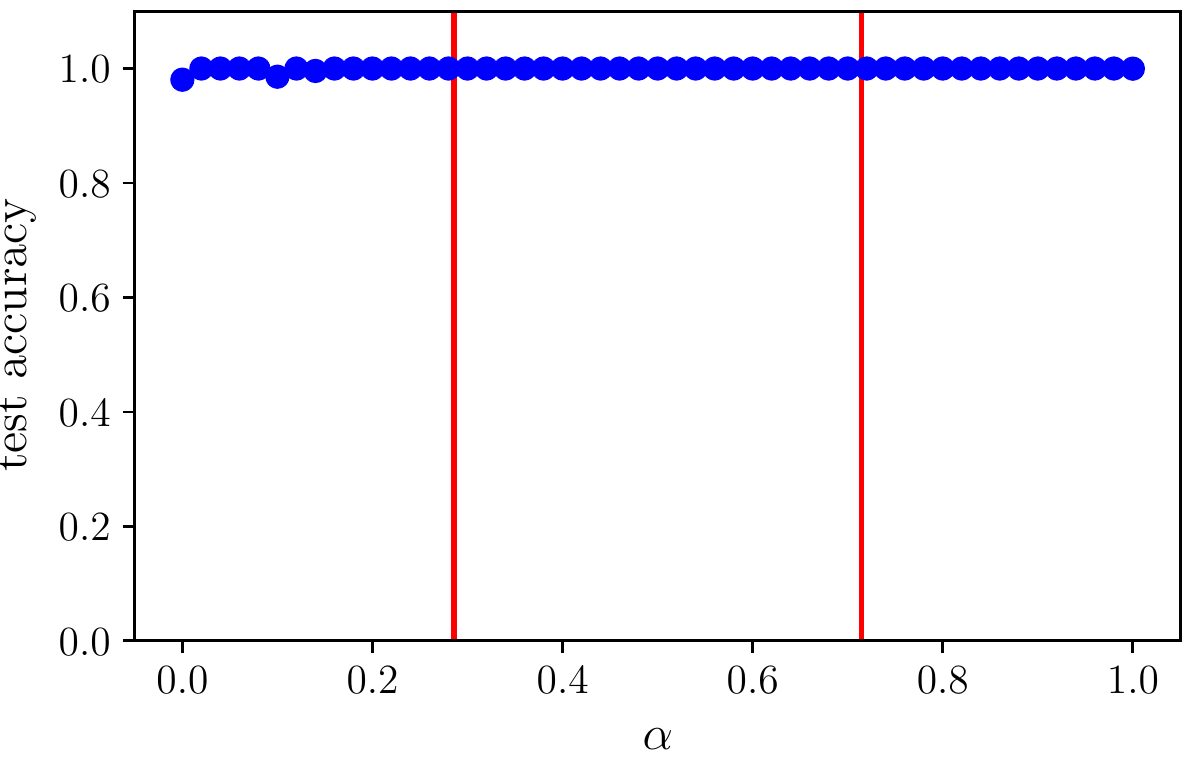}
		\caption{SE loss, with the correction}
	\end{subfigure}
	\caption{Testing accuracy with respect to $\alpha$ for Example 1 with and without the correction steps. The corruption ratio $\lambda=0.7$. Top row: trained with the CCE loss function; Bottom row: trained with the SE loss function. The blue dots represent the accuracy and the red lines represents the bounds of $\alpha$ in (\ref{ReCon2}).} 
	\label{fig:swiss_roll_bound}
\end{figure}

We first verify the theoretical condition \eqref{ReCon2} for the
complete recovery for binary classification problems. To this end, we
take the corruption probability $\alpha=0.02 k$ for $k=0, 1, ...,
50$. For each value of $\alpha$ we corrupt the labels and train the
neural network for a sufficiently large number of epochs such that the
training loss and the test accuracy attain a steady state. The
accuracy is tested on a separate clean test data set of size
$5,000$. We present the test accuracy for different values of $\alpha$
in \figref{fig:swiss_roll_bound}. To show the effectiveness of the
correction algorithm, we present the accuracy plot generated by the
classifier without the correction step as well. For the corruption
ratio $\lambda=0.7$, the classification can be fully recovered if
$\frac27 <\alpha< \frac57$, based on \eqref{ReCon2}. This inverval is
indicated by two red vertical lines in \figref{fig:swiss_roll_bound}
%
We clearly observe that when the corruption parameter $\lambda$ and
$\alpha$ satisfy the condition \eqref{ReCon2}, the classification can
be fully recovered with test accuracy almost $1$. Otherwise, the test
accuracy stays at around $0.5$, no matter how sufficient the training
is. Once our proposed correction algorithm is applied, the
classification is fully recovered for any value of $\alpha$. This
verifies the theoretical results in Corollary \ref{cor:2d} and the
effectiveness of the correction algorithm.

 
\begin{figure}[!htb]
	\centering
	\begin{subfigure}[b]{0.49\textwidth}
		\centering
		\includegraphics[width=0.9\linewidth]{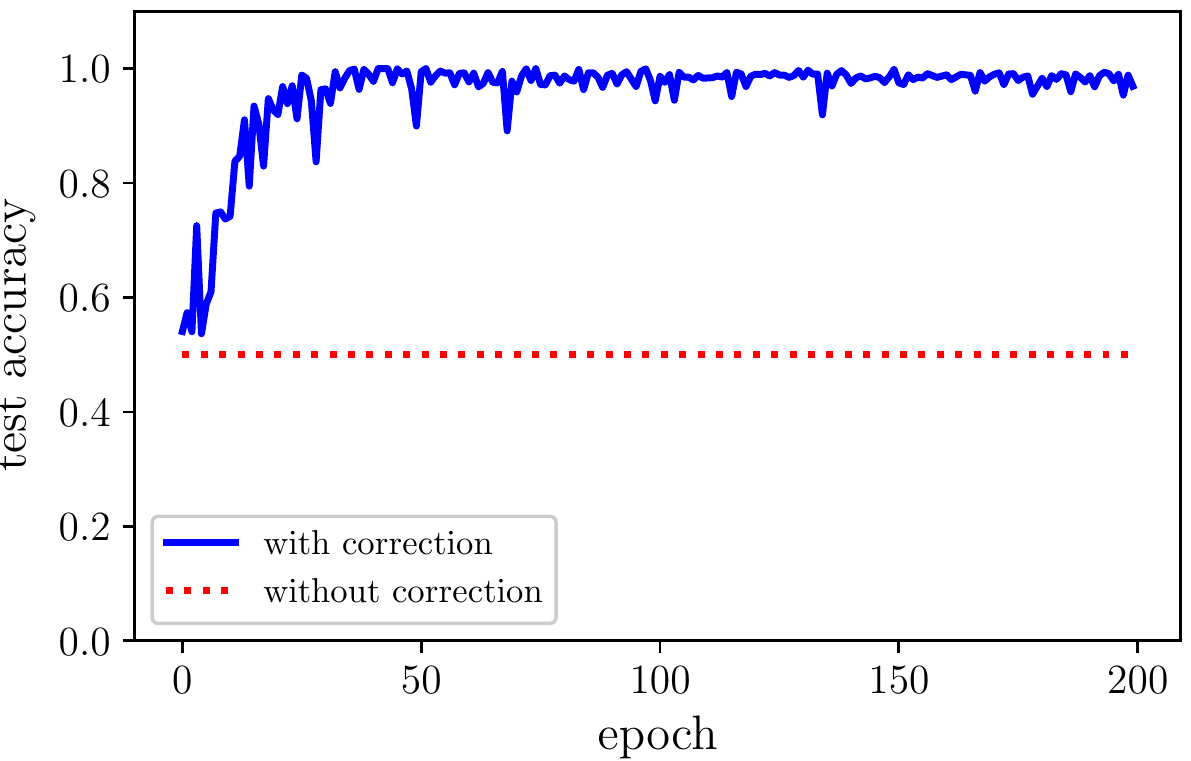}
		\caption{CCE loss, $\lambda=0.9$}
	\end{subfigure}
	\begin{subfigure}[b]{0.49\textwidth}
		\centering
		\includegraphics[width=0.9\linewidth]{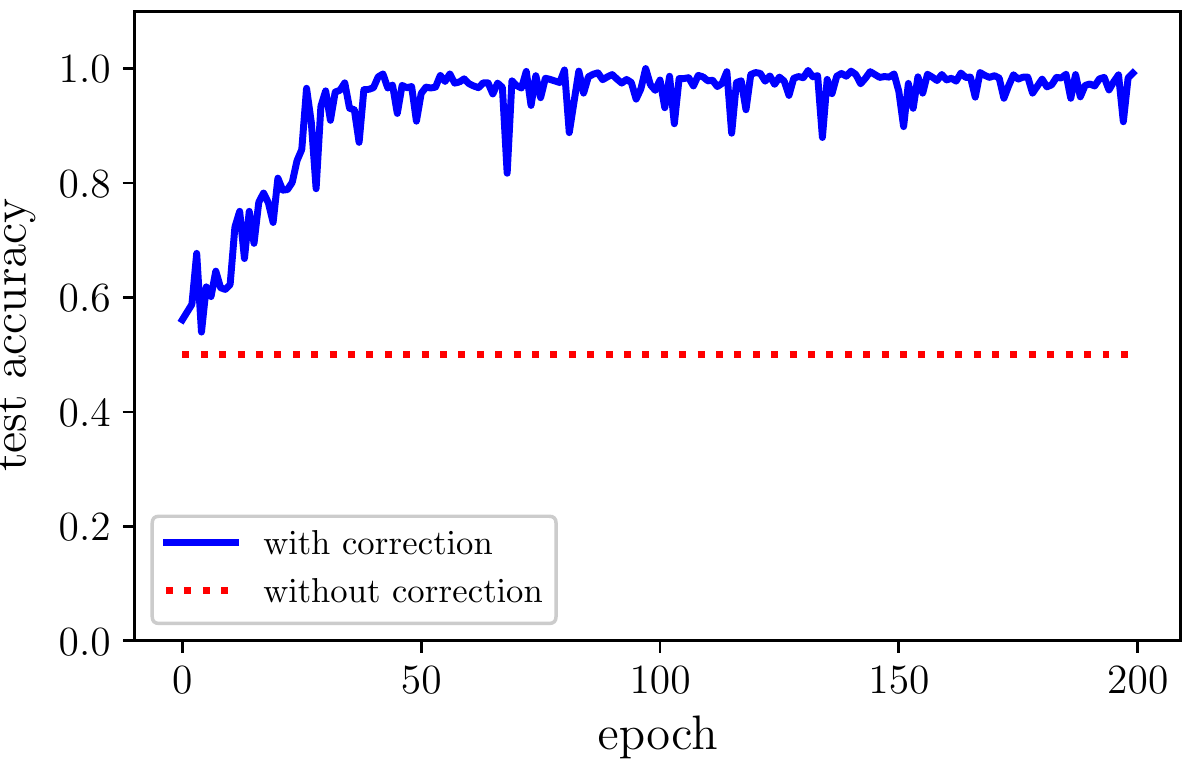}
		\caption{SE loss, $\lambda=0.9$}
	\end{subfigure}
	\caption{History of the testing accuracy during the training for Example 1 with corruption distribution $(0.3, 0.7)$ and the corruption ratio $\lambda=0.9$. The left is the training history with CCE loss function; The right is the training history with the SE loss function. The blue curve represents the testing accuracy with the correction step and the red curve is the result without the correction step.}
	\label{fig:swiss_roll_history} 
\end{figure}

\begin{figure}[!htb]
	\centering
	\begin{subfigure}[b]{0.49\textwidth}
		\centering
		\includegraphics[width=0.9\linewidth]{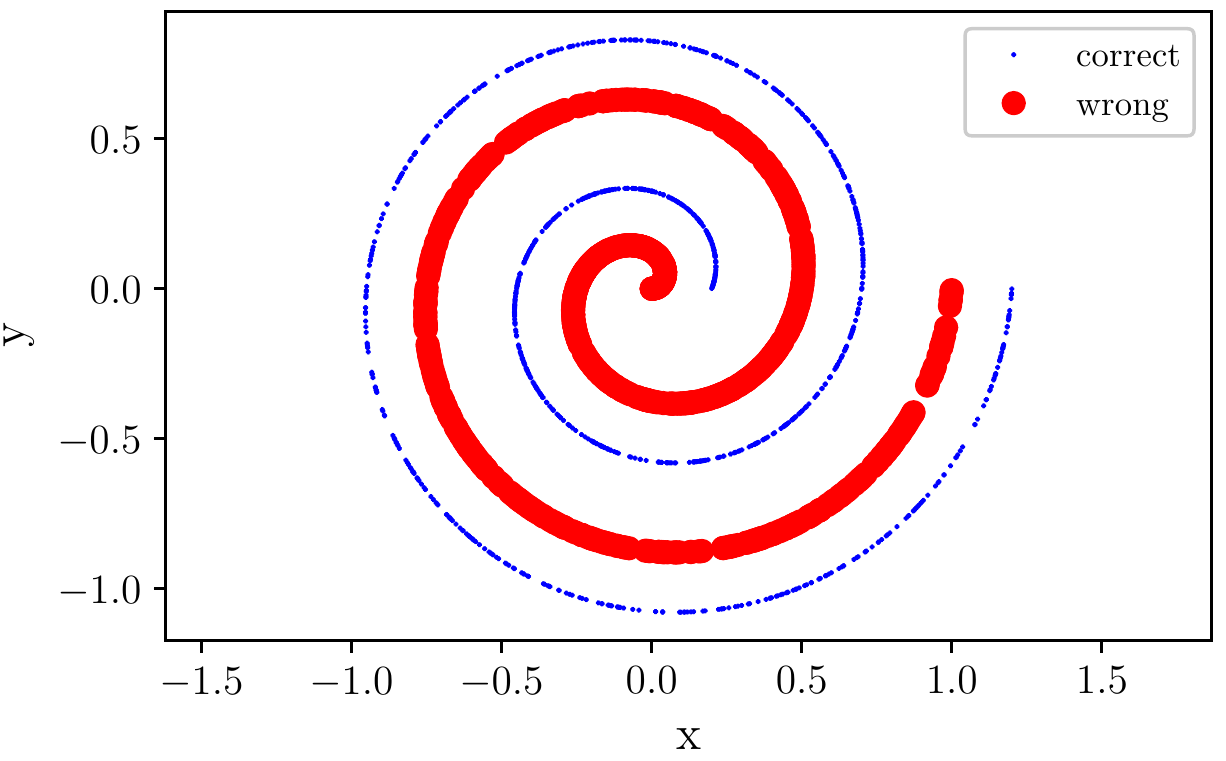}
		\caption{CCE loss, without the correction}
	\end{subfigure}
	\begin{subfigure}[b]{0.49\textwidth}
		\centering
		\includegraphics[width=0.9\linewidth]{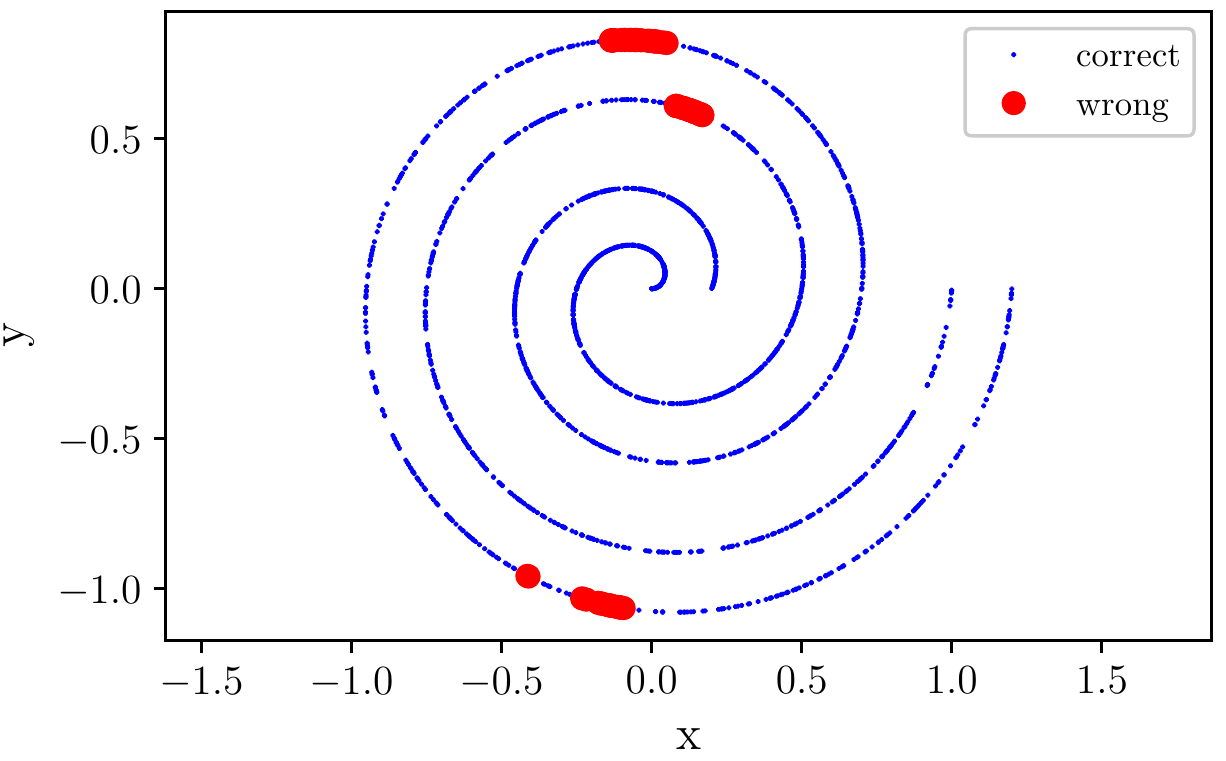}
		\caption{CCE loss, with the correction}
	\end{subfigure}
	\begin{subfigure}[b]{0.49\textwidth}
		\centering
		\includegraphics[width=0.9\linewidth]{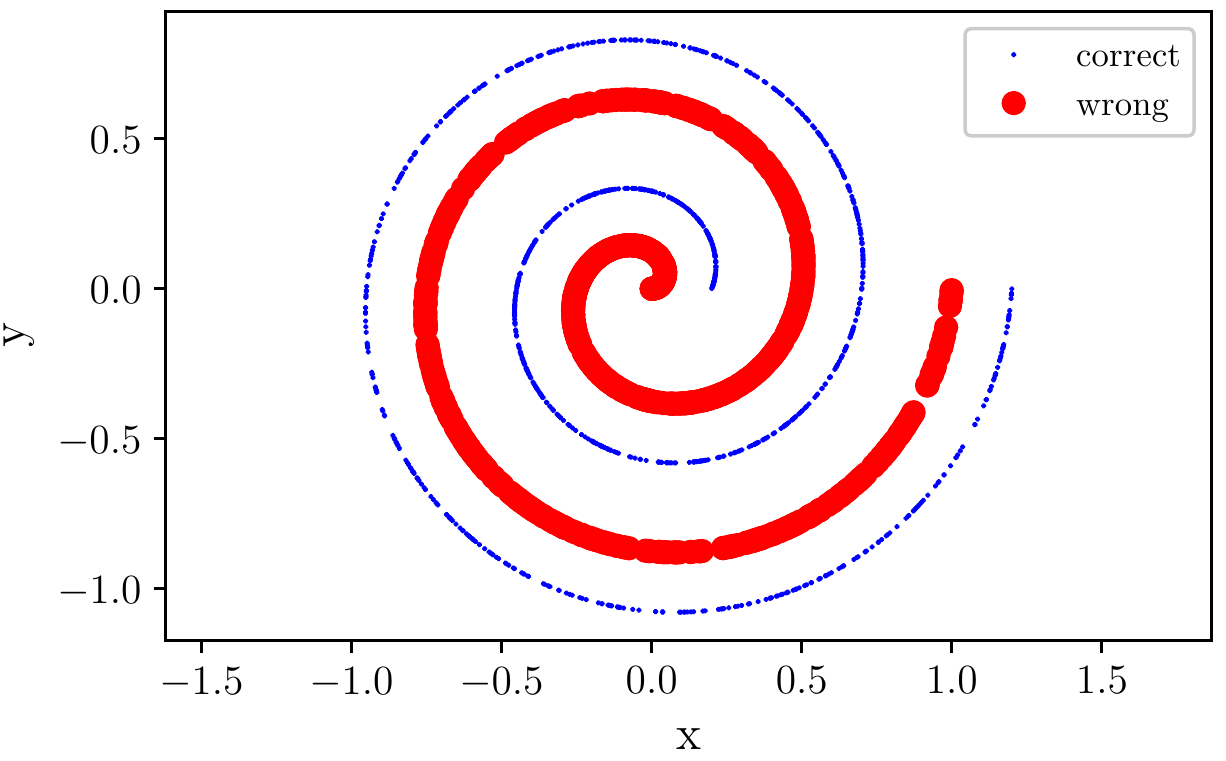}
		\caption{SE loss, without the correction}
	\end{subfigure}
	\begin{subfigure}[b]{0.49\textwidth}
		\centering
		\includegraphics[width=0.9\linewidth]{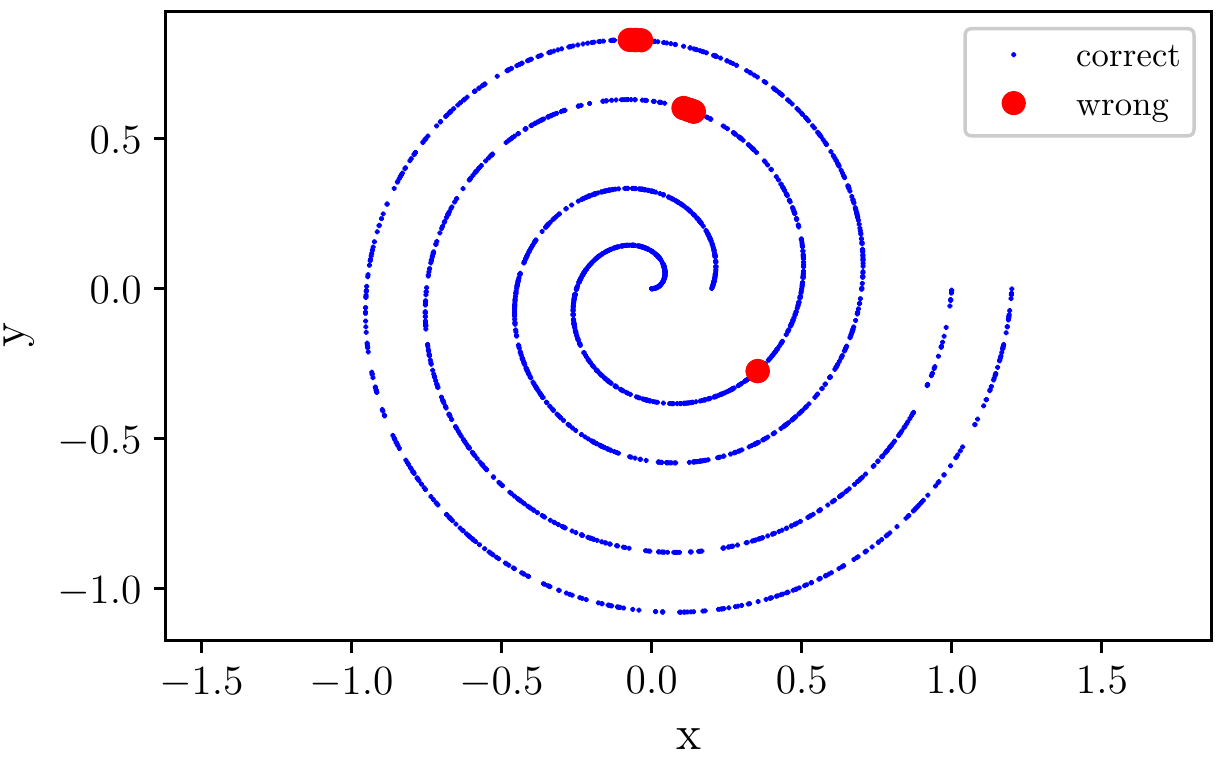}
		\caption{SE loss, with the correction}
	\end{subfigure}
	\caption{Prediction results on test data for Example 1 with and without the correction steps. The corruption ratio $\lambda=0.9$ and the corruption distribution is $(0.3, 0.7)$. Top row: trained with the CCE loss function; Bottom row: trained with the SE loss function. The blue dots represent the correct predictions and the red dots are the wrong ones.} 
	\label{fig:swiss_roll_prediction}
\end{figure}

To further verify the effectiveness of the correction algorithm. We
consider a more severe test with the corruption rate
$\lambda=0.9$. That is, 90\% of the data are corrupted. The corruption
probability is taken as $\alpha=0.3$, which does not satisfy the
condition \eqref{ReCon2}. We show the history of the test accuracy
during the training in \figref{fig:swiss_roll_history} for the cases
with and without the correction step. We see that in this severe test,
the correction algorithm can still attain a test accuracy near $1$,
whereas the standard neural network without the correction step
produce a test accuracy around $0.5$. This is due to the fact that
corruption is biased towards $D_2$. Most of the samples have the label
$\ve_2$ and hence the neural network tends to classify every feature
into the second class. This is illustrated by the prediction plot in
\ref{fig:swiss_roll_prediction}. With the correction algorithm applied, such
bias is eliminated and the classification becomes almost completely
recovered. 


\subsection*{Example 2: Multiple-class Classification}
\begin{figure}[!htb]
	\centering
	\includegraphics[scale=0.6]{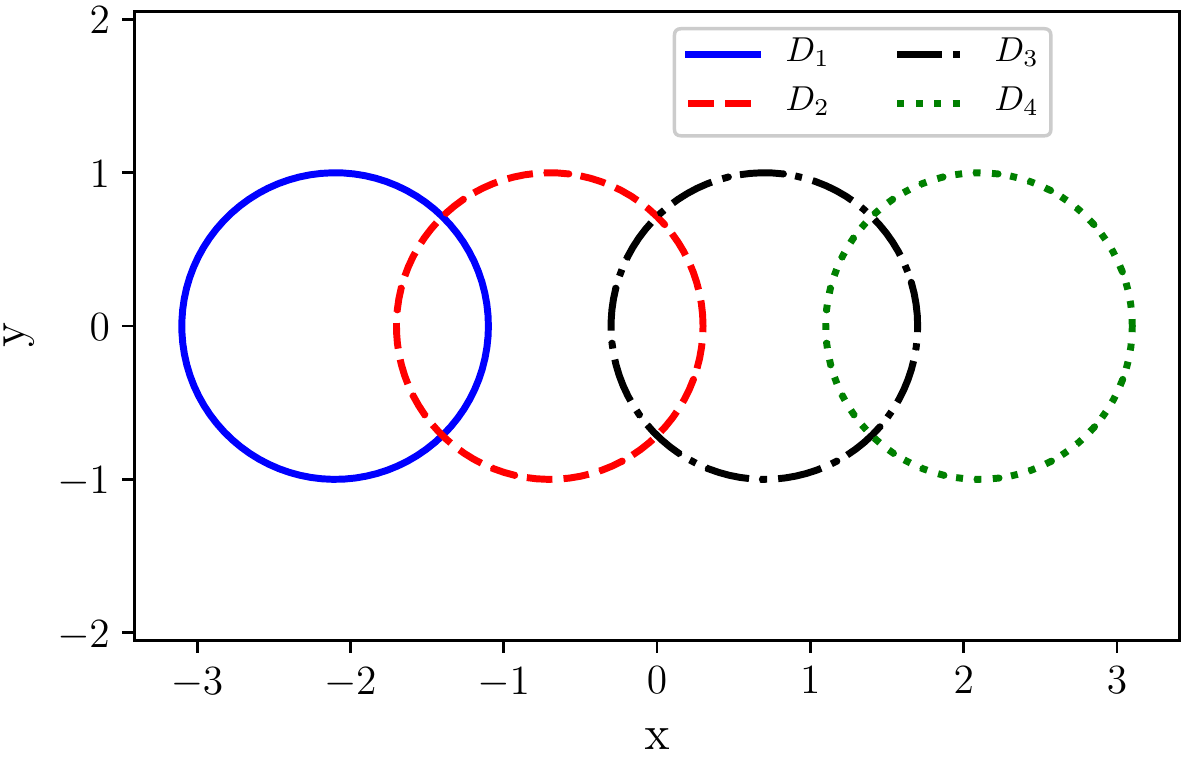}
	\caption{The illustration of the four-circle example.}
	\label{fig:four_circles}
\end{figure}

We further test our algorithm with multiple-class classification
problems. In this example we consider the classification of $n=4$
classes, consisting of four unit circles with centers at $(-2.1,\,
0)$, $(-0.7,\, 0)$, $(0.7,\, 0)$ and $(2.1,\,0)$ respectively, as
shown in \figref{fig:four_circles}. The training samples are drawn
from the uniform distribution over $D=\bigcup_{k=1}^4 D_k$. We take
$M=80,000$ samples with $20,000$ samples for each class. The labels are
corrupted with a corruption distribution probability of $(0.7, 0.1, 0.1, 0.1)$. For
the corruption ratio, we test two cases $\lambda=0.3$ and
$\lambda=0.7$. In the first case the condition
\eqref{ReCondition} is satisfied and the classification can be
completely recoverred without any correction, whereas in the second
case, the condition \eqref{ReCondition} is violated and a correction
step is necessary to recover the classification.

\begin{figure}[!htb]
	\centering
	\begin{subfigure}[b]{0.49\textwidth}
		\centering
		\includegraphics[width=0.9\linewidth]{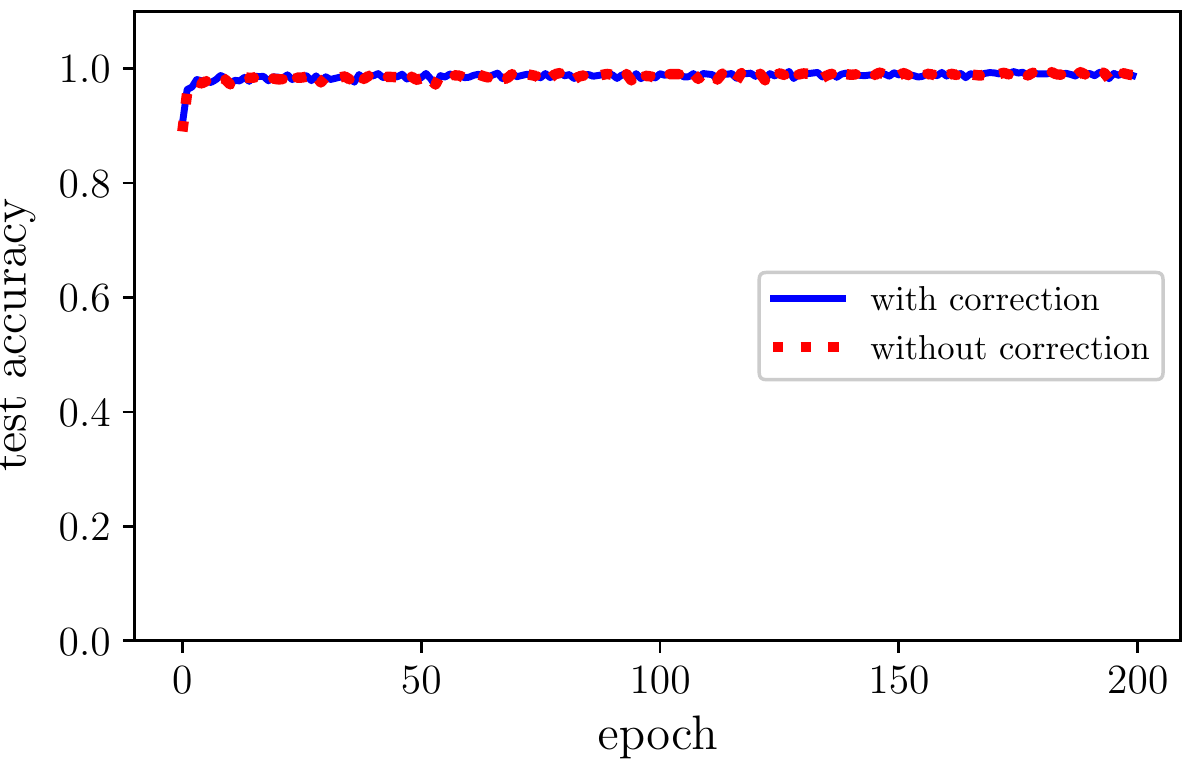}
		\caption{CCE loss, 30\% corruption }
	\end{subfigure}
	\begin{subfigure}[b]{0.49\textwidth}
		\centering
		\includegraphics[width=0.9\linewidth]{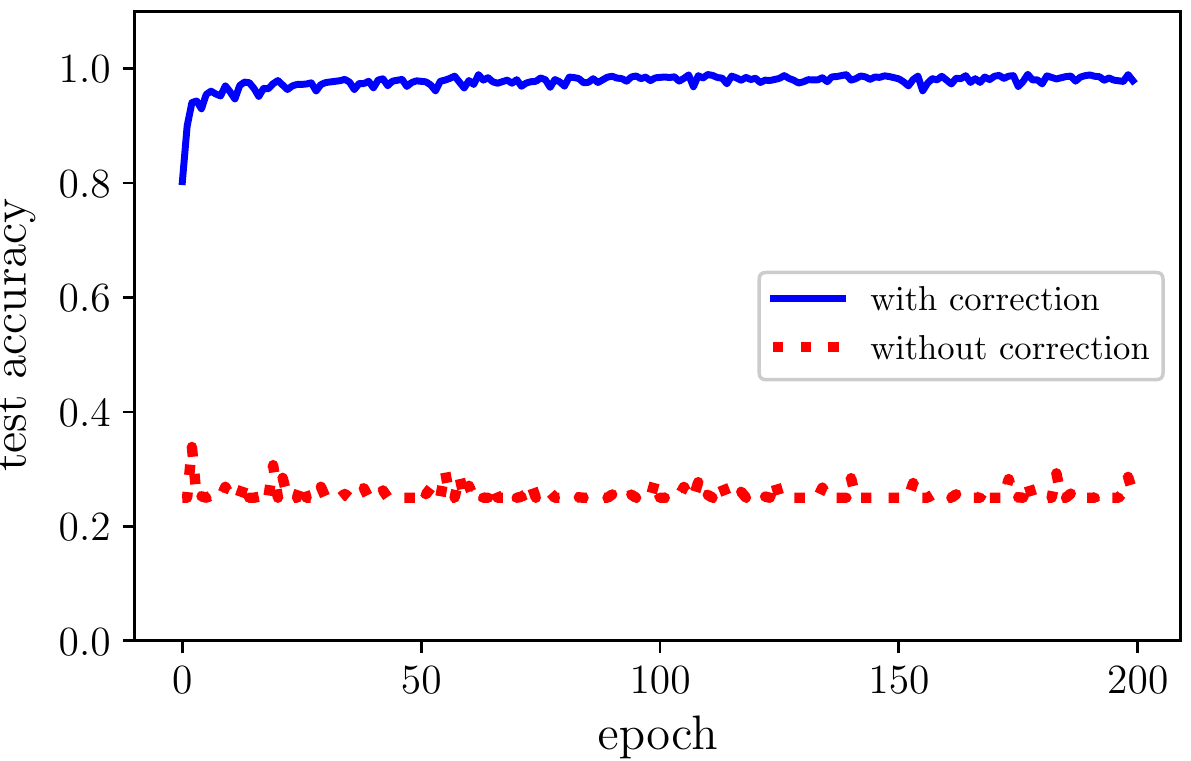}
		\caption{CCE loss, 70\% corruption }
	\end{subfigure}
	\begin{subfigure}[b]{0.49\textwidth}
		\centering
		\includegraphics[width=0.9\linewidth]{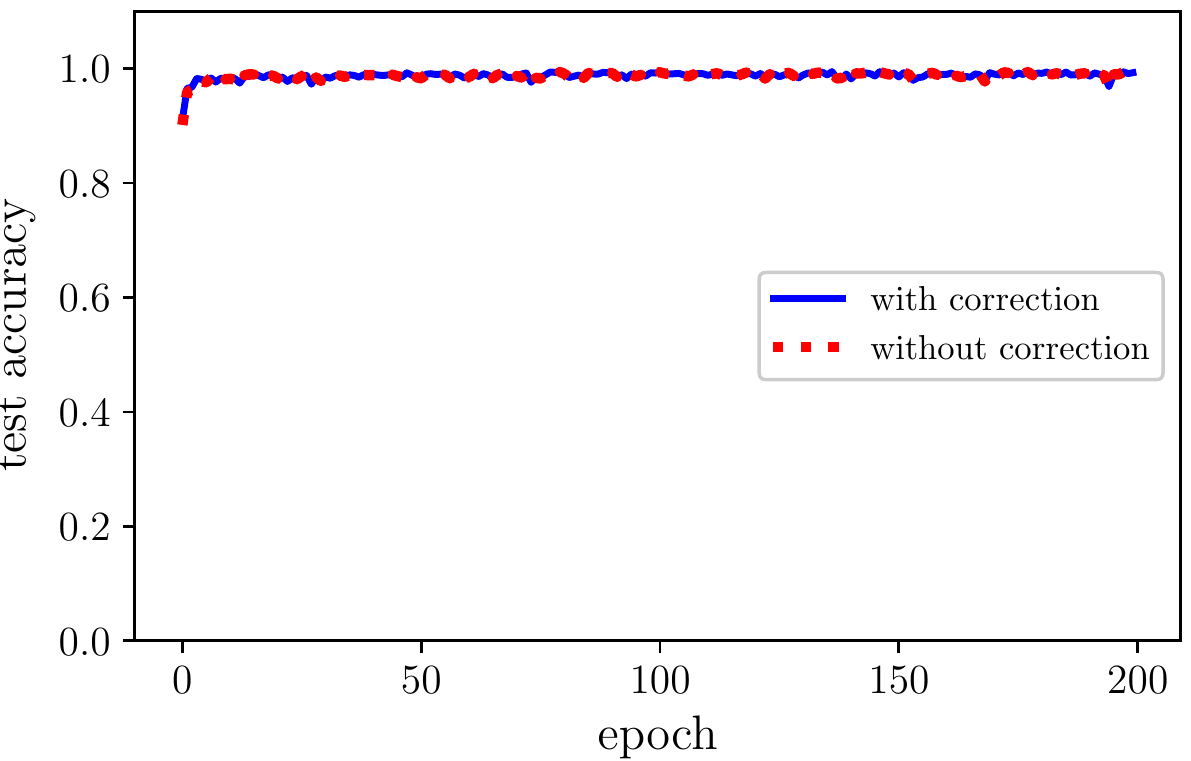}
		\caption{SE loss, 30\% corruption}
	\end{subfigure}
	\begin{subfigure}[b]{0.49\textwidth}
		\centering
		\includegraphics[width=0.9\linewidth]{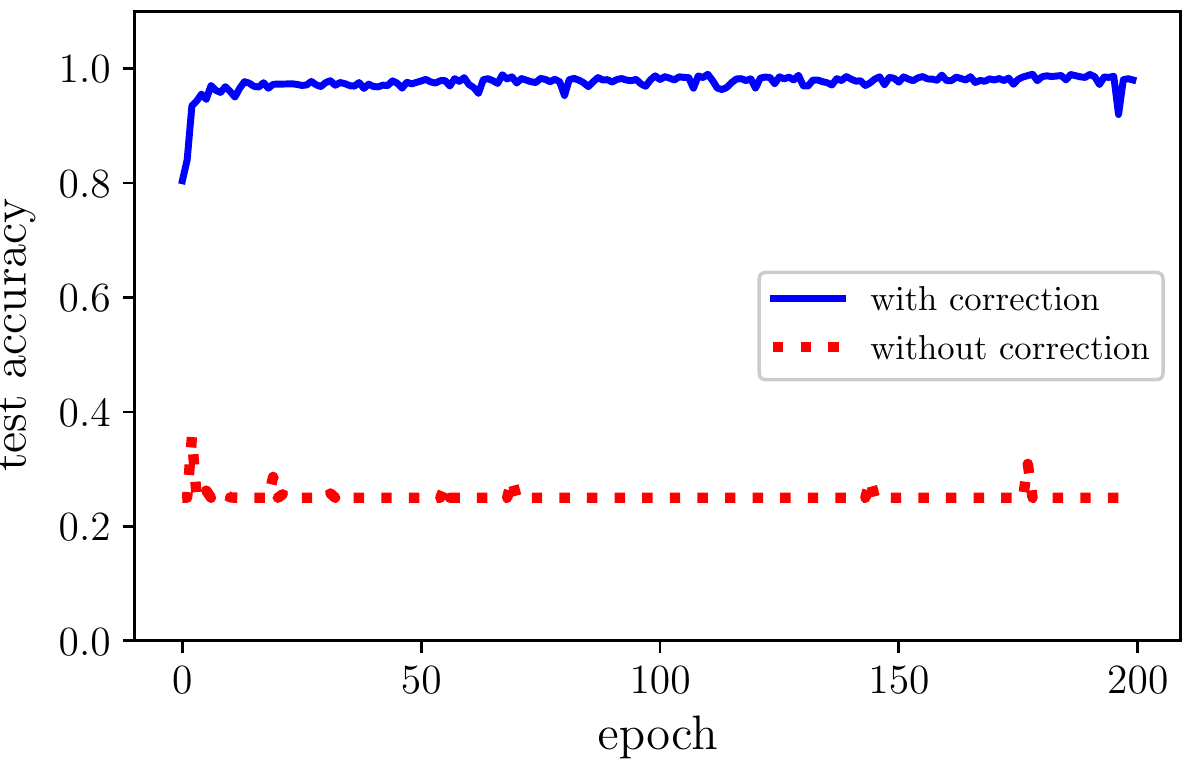}
		\caption{SE loss, 70\% corruption}
	\end{subfigure}

	\caption{History of the testing accuracy during the training for Example 2 with corruption distribution $(0.7, 0.1, 0.1, 0.1)$. Top row: the training history with CCE loss function; Bottom row: training history with the SE loss function. The left column is for the corruption ratio $\lambda=0.3$ and the right column is for $\lambda=0.7$. The blue curve represents the testing accuracy with the correction step and the red curve is the result without the correction step.}
	\label{fig:four_circle_history} 
\end{figure}

\begin{figure}[!htb]
	\centering
	\begin{subfigure}[b]{0.49\textwidth}
		\centering
		\includegraphics[width=0.9\linewidth]{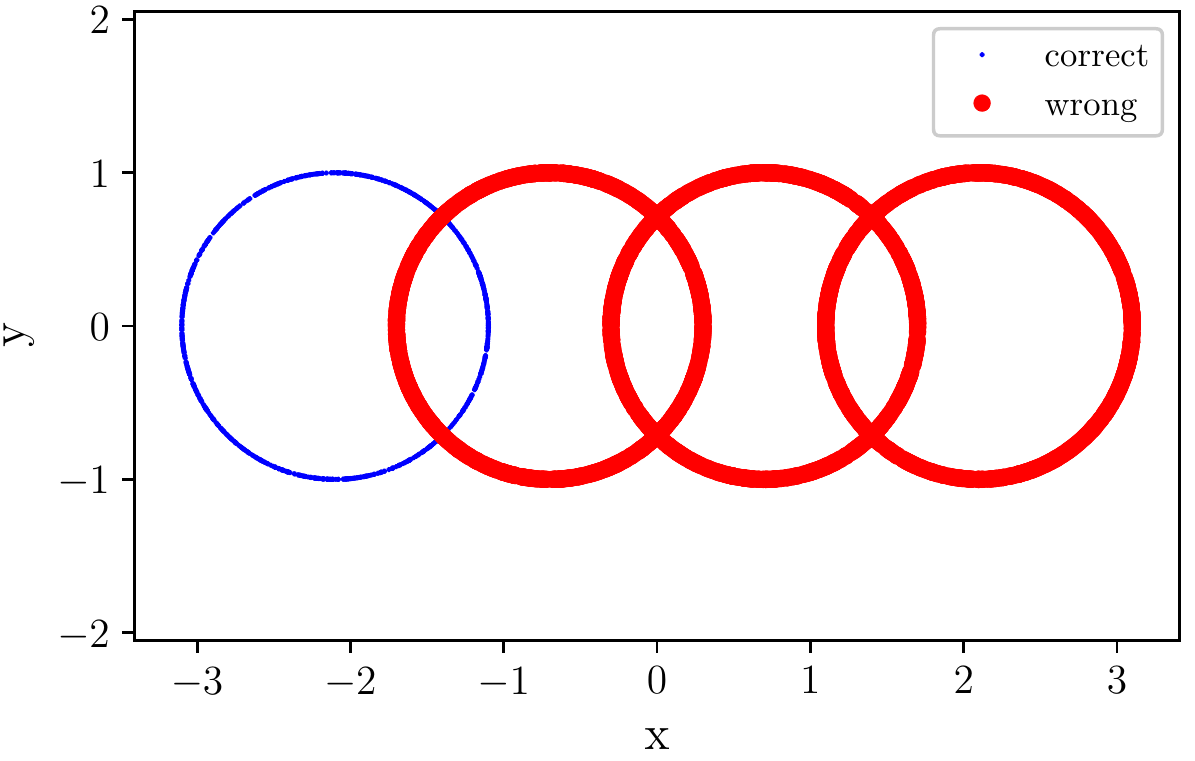}
		\caption{CCE loss, without the correction}
	\end{subfigure}
	\begin{subfigure}[b]{0.49\textwidth}
		\centering
		\includegraphics[width=0.9\linewidth]{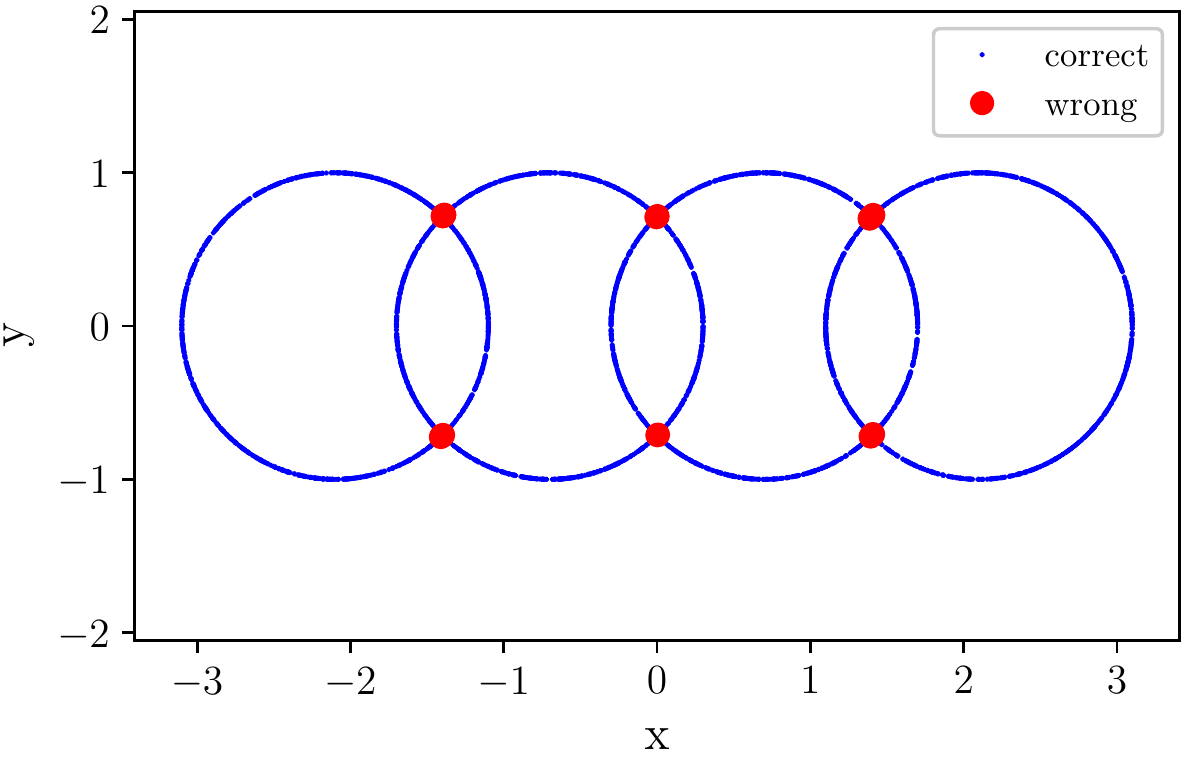}
		\caption{CCE loss, with the correction}
	\end{subfigure}
	\begin{subfigure}[b]{0.49\textwidth}
		\centering
		\includegraphics[width=0.9\linewidth]{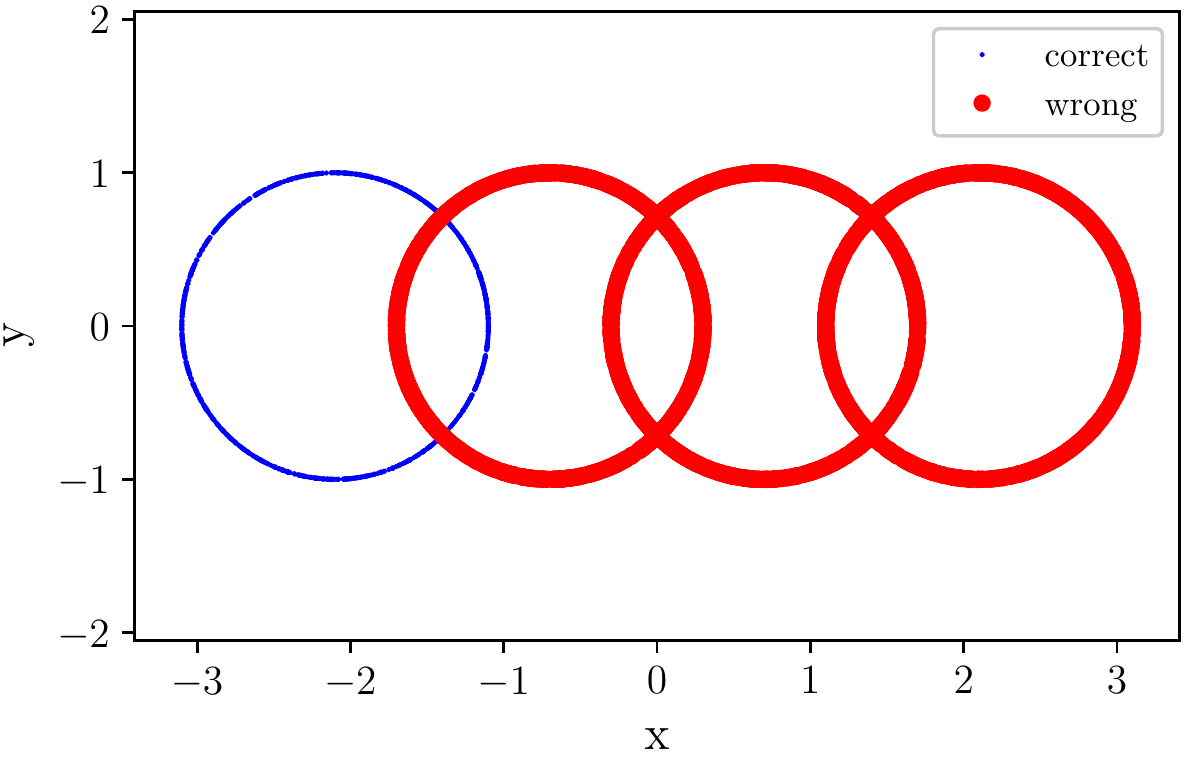}
		\caption{SE loss, without the correction}
	\end{subfigure}
	\begin{subfigure}[b]{0.49\textwidth}
		\centering
		\includegraphics[width=0.9\linewidth]{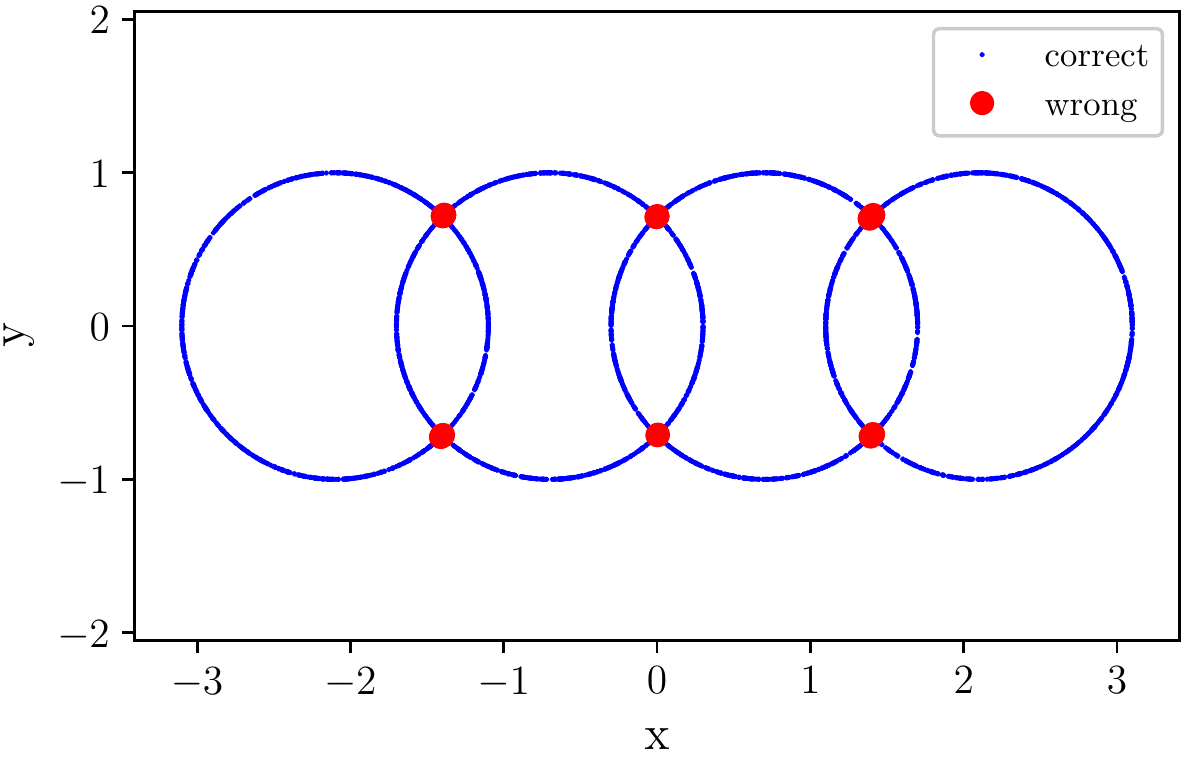}
		\caption{SE loss, with the correction}
	\end{subfigure}
	\caption{Prediction results on test data for Example 2 with and without the correction steps. The corruption ratio $\lambda=0.7$ and the corruption distribution is $(0.7, 0.1, 0.1, 0.1)$. Top row: trained with the CCE loss function; Bottom row: trained with the SE loss function. The blue dots represent the correct predictions and the red dots are the wrong ones.} 
	\label{fig:four_circle_prediction}
\end{figure}

We use the same neural network architecture as in Example 1. Both of
the CCE and SE loss functions are considered. The accuracy is tested
on a set of $2,000$ clean data. In \figref{fig:four_circle_history},
the history of the test accuracy during the training is presented for
both cases, with and without the correction step. It is observed that
for the case $\lambda=0.3<0.5$, the classification is always
completely recovered, regardless whether the correction step is added or
not. For the $\lambda=0.7$ case, where the condition \ref{ReCondition}
is not satisfied, the classifier without the correction produce a test
accuracy around $25\%$, whereas the correction step can recover the
classification and attain an accuracy almost $100\%$. This difference can
be more clearly observed in \figref{fig:four_circle_prediction}, where
the prediction results for the case $\lambda=0.7$ are presented. Since
the corruption distribution is biased towards the first class, the
trained classifier (without correction) predicts the first class
accurately but wrongly produces
predictions for the other three classes. The correction procedure helps
recover the classification, except for those not-well-defined samples that lie at the
intersections of two neighboring classes


\subsection*{Example 3: MINST Dataset}
Next, we test the applicability of the proposed correction algorithm
on classification of the MINST hand-written digits data set
\cite{lecun1990handwritten}, with corrupted labels. The labels are
corrupted with a fixed corruption distribution $(0.5, 0.04, 0.02,
0.03, 0.06, 0.07, 0.1, 0.08, 0.1, 0)$. By the condition
\eqref{ReCondition}, the classification can be completely recovered if
the corruption ratio $\lambda<\frac23\simeq 0.6667$. 

The classifier is constructed by using a convolutional neural network
(CNN), which consists of one 2D convolution layer of $32$ filters with
size $3\times 3$. It is followed by a max pooling layer and then the
other 2D convolution layer of $64$ filters. At last a dense layer with
$128$ nodes is added. To regularize the network, two dropout layers
with dropout rate $0.25$ and $0.5$ are added, both before and after the
dense layers. The activation functions are taken to be
ReLU except in the output layer, where the softmax function is
employed.  


The trained model is tested on $10,000$ clean testing data, and we
record the test accuracy for different corruption ratio $\lambda$ in
Table \ref{tab:MNIST_acc_cce} for the CCE loss and in Table
\ref{tab:MNIST_acc_se} for the SE loss. In both experiments, we record
the test accuracy both with and without correction, as well as 
the number of epochs used in the training. We clearly
observed the improvement in the test accuracy with the help of the
correction step, especially when the the corruption ratio is greater
than $0.5$. When the corruption ratio is $0.9$, the noise and
corruption in the dataset is so overwhelming, combined with the
limited number of training samples in MNIST,
that no amount of correction can help to recover the classification.
\begin{table}[!htb]
	\centering
	\begin{tabular}{c|c|c|c}
		\hline\hline
		corruption ratio & test accuracy & corrected test accuracy & \# epochs\\
		\hline\hline
		0\%& 0.9922& 0.9922& 12\\
		10\%& 0.9873& 0.9875& 10\\
		20\%& 0.9842& 0.9858& 8 \\
		30\%& 0.9781& 0.9836& 7\\
		40\%& 0.9412& 0.9745& 5\\
		50\%& 0.8655& 0.9710& 6\\
		60\%& 0.5051& 0.9607& 7\\
		70\%& 0.1294& 0.9331& 9\\
		80\%& 0.0981& 0.7875&10\\
		90\%& 0.0980& 0.0954& 4\\
		\hline
	\end{tabular}
	\caption{Test accuracy for the MNIST dataset with different corruption ratios and with and without the correction step. The model is trained with the CCE loss function.}
	\label{tab:MNIST_acc_cce}
\end{table}

\begin{table}[!htb]
	\centering
	\begin{tabular}{c|c|c|c}
		\hline\hline
		corruption ratio & test accuracy & corrected test accuracy & \# epochs\\
		\hline\hline
		0\%& 0.9914& 0.9914& 11\\
		10\%& 0.9897& 0.9898& 7\\
		20\%& 0.9868& 0.9886& 8\\
		30\%& 0.9755& 9.9799& 5\\
		40\%& 0.9529& 0.9765& 10\\
		50\%& 0.9480& 0.9712& 4\\
		60\%& 0.5599& 0.9590& 7\\
		70\%& 0.1207& 0.9330& 9\\
		80\%& 0.0998& 0.7692& 13\\
		90\%& 0.0980& 0.0958& 4\\
		\hline
	\end{tabular}
	\caption{Test accuracy for the MNIST dataset with different corruption ratios and with and without the correction. The model is trained with the SE loss function.}
	\label{tab:MNIST_acc_se}
\end{table}

\subsection*{Example 4: Fashion MNIST Dataset}
Our last example is the fashion MNIST dataset \cite{xiao2017fashion},
consisting of $28\times 28$ gray-scale images of fashion products from
$10$ categories. The data set has $70,000$ samples in total. We take $60,000$ as training data and the other $10,000$ as testing data.

The labels of the training samples are corrupted according to a fixed
corruption distribution $(0.5, 0.04, 0.02, 0.03, 0.06, 0.07, 0.1,
0.08, 0.1, 0)$ with different corruption ratios.
%
We use the same neural network structure as that for  Example 3.
In Table \ref{tab:Fashion_MNIST_acc_cce} and Table
\ref{tab:Fashion_MNIST_acc_se}, we record the test accuracy for the
model trained with the CCE and SE loss functions, both before and
after applying the correction algorithm. Again,
significant improvement in accuracy is observed, except the
extreme case where the corruption ratio is $90\%$.

\begin{table}[!htb]
	\centering
	\begin{tabular}{c|c|c|c}
		\hline\hline
		corruption ratio & test accuracy & corrected test accuracy & \# epochs \\
		\hline\hline
		0 \%& 0.9204& 0.9204& 10\\
		10\%& 0.9080& 0.9080& 7\\
		20\%& 0.9098& 0.9106& 9\\
		30\%& 0.8975& 0.9013& 10\\
		40\%& 0.8770& 0.8928& 8\\
		50\%& 0.7880& 0.8798& 11\\
		60\%& 0.5947& 0.8771& 7\\
		70\%& 0.1374& 0.8421& 9\\
		80\%& 0.1001& 0.7397& 13\\
		90\%& 0.1000& 0.1000& 4\\
		\hline
	\end{tabular}
	\caption{Test accuracy for the Fashion MNIST dataset with different corruption ratios and with and without the correction. The model is trained with the CCE loss function.}
	\label{tab:Fashion_MNIST_acc_cce}
\end{table}

\begin{table}[!htb]
	\centering
	\begin{tabular}{c|c|c|c}
		\hline\hline
		corruption ratio & test accuracy & corrected test accuracy & \# epochs\\
		\hline\hline
		0\%& 0.9193& 0.9193& 10\\
		10\%& 0.9148& 0.9145& 7\\
		20\%& 0.9057& 0.9069& 14\\
		30\%& 0.8937& 0.8951& 5\\
		40\%& 0.8806& 0.8879& 6\\
		50\%& 0.8276& 0.8804& 8\\
		60\%& 0.5792& 0.8735& 7\\
		70\%& 0.1166& 0.8375& 10\\
		80\%& 0.1000& 0.7519& 13\\
		90\%& 0.1000& 0.1000& 7\\
		\hline
	\end{tabular}
	\caption{Test accuracy for the Fashion MNIST dataset with different corruption ratios and with and without the correction. The model is trained with the SE loss function.}
	\label{tab:Fashion_MNIST_acc_se}
\end{table}

%% file: Conclusion.tex
\section{Conclusion} \label{sec:conclusions}

In this paper, we proposed a correction algorithm for the classification problems, where the data available are potentially corrupted. When the model is trained by minimizing the CCE or SE loss function, given sufficiently large amount of data, it is theoretically shown that the classification can be completely recovered by adding a correction step to the trained model. In particular, if the training data contains unlabeled samples, a random label assignment according to the uniform distribution would make the classification completely recovered. The proposed algorithm is non-intrusive and can be coupled with many models, such as support vector machines, neural networks of various architectures and so on. 
Numerical experiments were conducted with the proposed correction procedure applied to the neural networks models for two academic examples as well as two benchmark real-world tests with complicated and limited dataset. 
Numerical results confirmed
the theoretical findings and demonstrated that, when the data labels contain corruptions, the proposed correction algorithm 
 gives satisfactory test accuracy and can effectively eliminate the impact of label corruptions.

%% file: Appendix.tex
\appendix

\section{Proof of Theorem \ref{thm:nd}}\label{app1:proof}
\begin{proof}
Note that 
$$
\min_{{\f}} J(\f) = \sum_{j=1}^n \min_\f  J_{j} (\f),
$$
where 
$$
J_{j} (\f) :=
\int_{D_j} \bigg( (1-\lambda) L ( {\bf e}_j, \f (\x) )
+  \lambda  \sum_{k=1}^n \alpha_k 
L ( {\bf e}_k, \f (\x) )
\bigg ) d \omega_j.
$$
For each $j=1,\dots,n$, we consider the minimum of the function
$$
F_j( f_1, f_2,\dots,f_n)=
(1-\lambda) L ( {\bf e}_j, \f  )
+  \lambda  \sum_{k=1}^n \alpha_k 
L ( {\bf e}_k, \f  ),
$$
subject to 
$$
\f = (f_1,f_2,\dots, f_n) \in \C,
$$
where $\C$ is the probability simplex.

For the CCE loss function, we have 
$$
F_j( f_1, f_2,\dots,f_n)=
-( 1-\lambda ) \log f_j 
-  \lambda  \sum_{k=1}^n \alpha_k \log f_k. 
$$
Using the Lagrangian multiplier method, we consider the Lagrange function
$$
F_j( f_1, f_2,\dots, f_n) +t \bigg( \sum_{k=1}^n f_k - 1\bigg).
$$
At the minimal point, the gradient of the Lagrange function must be zero. This yields
\begin{equation}\label{key22}
\begin{cases}
\frac{\lambda a_k}{ f_k} -  t = 0, \quad \forall k\neq j,
\\
\frac{1-\lambda}{ f_j} + \frac{\lambda a_j}{ f_j} -  t = 0,
\\
\sum_{k=1}^n f_k = 1.
\end{cases}
\end{equation}
Solving \eqref{key22}, one obtains 
$$
\begin{cases}
f_j = 1-\lambda + \lambda \alpha_{j},
\\
f_k = \lambda \alpha_k, \quad \forall k\neq j.
\end{cases}	
$$
Hence, the function $\f (\x)$ that minimizes $J(\f)$ satisfies: 
$$
\f (\x) = \left( \lambda \alpha_1, \cdots, \lambda \alpha_{j-1}, 
{1-\lambda}+\lambda \alpha_{j},
\cdots, 
{\lambda \alpha_{j+1}}, \cdots,
{\lambda \alpha_{n}}
\right), \quad \x \in D_j,
$$
for $1\le j \le n$.

For the SE loss function, we have 
\begin{align*}
F_j( f_1, f_2,\dots, f_n) & =
(1-\lambda) \left( \sum_{\ell \neq j} f_\ell^2 + ( f_j-1)^2  \right) 
+ \lambda \sum_{k=1}^n \alpha_k 
\left( \sum_{\ell \neq k} f_\ell^2 + ( f_k-1)^2 \right)
\\
& = (1-\lambda + \lambda \alpha_j  ) \left( \sum_{\ell \neq j} f_\ell^2 + ( f_j-1)^2  \right) 
+ \lambda \sum_{k \ne j } \alpha_k 
\left( \sum_{\ell \neq k} f_\ell^2 + ( f_k-1)^2 \right)
\end{align*}
Using the Lagrangian multiplier method, we consider the Lagrange function
$$
F_j( f_1, f_2,\cdots, f_n) +t \bigg( \sum_{k=1}^n f_k - 1\bigg).
$$
At the minimal point, the gradient of the Lagrange function must be zero. This yields
\begin{equation*}
\begin{cases}
2(1-\lambda + \lambda \alpha_j) f_k + 2 \lambda \alpha_k (f_k-1) 
+ 2 \lambda f_k \sum_{ i \ne k, i \ne j } \alpha_i +  t = 0, \quad \forall k\neq j,
\\
2 ( 1-\lambda + \lambda \alpha_j ) (f_j-1)  + 2 \lambda f_j \sum_{k\ne j} \alpha_k +  t = 0,
\\
\sum_{k=1}^n f_k = 1,
\end{cases}
\end{equation*}	
which further implies 
\begin{equation}\label{key223} 
\begin{cases}
f_k = \lambda \alpha_k - \frac{t}2, \quad \forall k\neq j,
\\
f_j = 1-\lambda + \lambda \alpha_j -\frac{t}2,
\\
\sum_{k=1}^n f_k = 1,
\end{cases}
\end{equation}		
Solving \eqref{key223}, one obtains 
$$
\begin{cases}
f_j = 1-\lambda + \lambda \alpha_{j},
\\
f_k = \lambda \alpha_k, \quad \forall k\neq j.
\end{cases}	
$$
Hence, the function $\f (\x)$ that minimizes $J(\f)$ satisfies: 
$$
\f (\x) = \left( \lambda \alpha_1, \cdots, \lambda \alpha_{j-1}, 
{1-\lambda}+\lambda \alpha_{j},
\cdots, 
{\lambda \alpha_{j+1}}, \cdots,
{\lambda \alpha_{n}}
\right), \quad \x \in D_j,
$$
for $1\le j \le n$.

The proof is complete.
\end{proof}

%% file: main.bbl
\begin{thebibliography}{1}

\bibitem{haber2017stable}
{\sc E.~Haber and L.~Ruthotto}, {\em Stable architectures for deep neural
  networks}, Inverse Problems, 34 (2017), p.~014004.

\bibitem{kingma2014adam}
{\sc D.~P. Kingma and J.~Ba}, {\em Adam: A method for stochastic optimization},
  arXiv preprint arXiv:1412.6980,  (2014).

\bibitem{lecun1990handwritten}
{\sc Y.~LeCun, B.~E. Boser, J.~S. Denker, D.~Henderson, R.~E. Howard, W.~E.
  Hubbard, and L.~D. Jackel}, {\em Handwritten digit recognition with a
  back-propagation network}, in Advances in neural information processing
  systems, 1990, pp.~396--404.

\bibitem{nair2010rectified}
{\sc V.~Nair and G.~E. Hinton}, {\em Rectified linear units improve restricted
  boltzmann machines}, in Proceedings of the 27th international conference on
  machine learning (ICML-10), 2010, pp.~807--814.

\end{thebibliography}


\begin{thebibliography}{10}

\bibitem{tensorflow2015-whitepaper}
{\sc M.~Abadi, A.~Agarwal, P.~Barham, E.~Brevdo, Z.~Chen, C.~Citro, G.~S.
  Corrado, A.~Davis, J.~Dean, M.~Devin, S.~Ghemawat, I.~Goodfellow, A.~Harp,
  G.~Irving, M.~Isard, Y.~Jia, R.~Jozefowicz, L.~Kaiser, M.~Kudlur,
  J.~Levenberg, D.~Man\'{e}, R.~Monga, S.~Moore, D.~Murray, C.~Olah,
  M.~Schuster, J.~Shlens, B.~Steiner, I.~Sutskever, K.~Talwar, P.~Tucker,
  V.~Vanhoucke, V.~Vasudevan, F.~Vi\'{e}gas, O.~Vinyals, P.~Warden,
  M.~Wattenberg, M.~Wicke, Y.~Yu, and X.~Zheng}, {\em {TensorFlow}: Large-scale
  machine learning on heterogeneous systems}, 2015,
  \url{http://tensorflow.org/}.
\newblock Software available from tensorflow.org.

\bibitem{brooks2011support}
{\sc J.~P. Brooks}, {\em Support vector machines with the ramp loss and the
  hard margin loss}, Operations research, 59 (2011), pp.~467--479.

\bibitem{chollet2015keras}
{\sc F.~Chollet et~al.}, {\em Keras}.
\newblock \url{https://keras.io}, 2015.

\bibitem{frenay2014classification}
{\sc B.~Fr{\'e}nay and M.~Verleysen}, {\em Classification in the presence of
  label noise: a survey}, IEEE transactions on neural networks and learning
  systems, 25 (2014), pp.~845--869.

\bibitem{ghosh2017robust}
{\sc A.~Ghosh, H.~Kumar, and P.~Sastry}, {\em Robust loss functions under label
  noise for deep neural networks}, in Thirty-First AAAI Conference on
  Artificial Intelligence, 2017.

\bibitem{ghosh2015making}
{\sc A.~Ghosh, N.~Manwani, and P.~Sastry}, {\em Making risk minimization
  tolerant to label noise}, Neurocomputing, 160 (2015), pp.~93--107.

\bibitem{graves2013speech}
{\sc A.~Graves, A.-r. Mohamed, and G.~Hinton}, {\em Speech recognition with
  deep recurrent neural networks}, in Acoustics, speech and signal processing
  (icassp), 2013 ieee international conference on, IEEE, 2013, pp.~6645--6649.

\bibitem{haber2017stable}
{\sc E.~Haber and L.~Ruthotto}, {\em Stable architectures for deep neural
  networks}, Inverse Problems, 34 (2017), p.~014004.

\bibitem{he2016deep}
{\sc K.~He, X.~Zhang, S.~Ren, and J.~Sun}, {\em Deep residual learning for
  image recognition}, in Proceedings of the IEEE conference on computer vision
  and pattern recognition, 2016, pp.~770--778.

\bibitem{hendrycks2018using}
{\sc D.~Hendrycks, M.~Mazeika, D.~Wilson, and K.~Gimpel}, {\em Using trusted
  data to train deep networks on labels corrupted by severe noise}, in Advances
  in Neural Information Processing Systems, 2018, pp.~10477--10486.

\bibitem{jiang2017mentornet}
{\sc L.~Jiang, Z.~Zhou, T.~Leung, L.-J. Li, and L.~Fei-Fei}, {\em Mentornet:
  Regularizing very deep neural networks on corrupted labels}, arXiv preprint
  arXiv:1712.05055, 4 (2017).

\bibitem{khetan2017learning}
{\sc A.~Khetan, Z.~C. Lipton, and A.~Anandkumar}, {\em Learning from noisy
  singly-labeled data}, arXiv preprint arXiv:1712.04577,  (2017).

\bibitem{kingma2014adam}
{\sc D.~P. Kingma and J.~Ba}, {\em Adam: A method for stochastic optimization},
  arXiv preprint arXiv:1412.6980,  (2014).

\bibitem{larsen1998design}
{\sc J.~Larsen, L.~Nonboe, M.~Hintz-Madsen, and L.~K. Hansen}, {\em Design of
  robust neural network classifiers}, in Proceedings of the 1998 IEEE
  International Conference on Acoustics, Speech and Signal Processing,
  ICASSP'98 (Cat. No. 98CH36181), vol.~2, IEEE, 1998, pp.~1205--1208.

\bibitem{lecun1990handwritten}
{\sc Y.~LeCun, B.~E. Boser, J.~S. Denker, D.~Henderson, R.~E. Howard, W.~E.
  Hubbard, and L.~D. Jackel}, {\em Handwritten digit recognition with a
  back-propagation network}, in Advances in neural information processing
  systems, 1990, pp.~396--404.

\bibitem{li2016data}
{\sc B.~Li, Y.~Wang, A.~Singh, and Y.~Vorobeychik}, {\em Data poisoning attacks
  on factorization-based collaborative filtering}, in Advances in neural
  information processing systems, 2016, pp.~1885--1893.

\bibitem{li2017learning}
{\sc Y.~Li, J.~Yang, Y.~Song, L.~Cao, J.~Luo, and L.-J. Li}, {\em Learning from
  noisy labels with distillation}, in Proceedings of the IEEE International
  Conference on Computer Vision, 2017, pp.~1910--1918.

\bibitem{liu2016classification}
{\sc T.~Liu and D.~Tao}, {\em Classification with noisy labels by importance
  reweighting}, IEEE Transactions on pattern analysis and machine intelligence,
  38 (2016), pp.~447--461.

\bibitem{long2010random}
{\sc P.~M. Long and R.~A. Servedio}, {\em Random classification noise defeats
  all convex potential boosters}, Machine learning, 78 (2010), pp.~287--304.

\bibitem{manwani2013noise}
{\sc N.~Manwani and P.~Sastry}, {\em Noise tolerance under risk minimization},
  IEEE transactions on cybernetics, 43 (2013), pp.~1146--1151.

\bibitem{masnadi2009design}
{\sc H.~Masnadi-Shirazi and N.~Vasconcelos}, {\em On the design of loss
  functions for classification: theory, robustness to outliers, and
  savageboost}, in Advances in neural information processing systems, 2009,
  pp.~1049--1056.

\bibitem{menon2015learning}
{\sc A.~Menon, B.~Van~Rooyen, C.~S. Ong, and B.~Williamson}, {\em Learning from
  corrupted binary labels via class-probability estimation}, in International
  Conference on Machine Learning, 2015, pp.~125--134.

\bibitem{mnih2012learning}
{\sc V.~Mnih and G.~E. Hinton}, {\em Learning to label aerial images from noisy
  data}, in Proceedings of the 29th International conference on machine
  learning (ICML-12), 2012, pp.~567--574.

\bibitem{nair2010rectified}
{\sc V.~Nair and G.~E. Hinton}, {\em Rectified linear units improve restricted
  boltzmann machines}, in Proceedings of the 27th international conference on
  machine learning (ICML-10), 2010, pp.~807--814.

\bibitem{natarajan2013learning}
{\sc N.~Natarajan, I.~S. Dhillon, P.~K. Ravikumar, and A.~Tewari}, {\em
  Learning with noisy labels}, in Advances in neural information processing
  systems, 2013, pp.~1196--1204.

\bibitem{Nettleton2010}
{\sc D.~F. Nettleton, A.~Orriols-Puig, and A.~Fornells}, {\em A study of the
  effect of different types of noise on the precision of supervised learning
  techniques}, Artificial Intelligence Review, 33 (2010), pp.~275--306.

\bibitem{northcutt2017learning}
{\sc C.~G. Northcutt, T.~Wu, and I.~L. Chuang}, {\em Learning with confident
  examples: Rank pruning for robust classification with noisy labels}, arXiv
  preprint arXiv:1705.01936,  (2017).

\bibitem{patrini2017making}
{\sc G.~Patrini, A.~Rozza, A.~Krishna~Menon, R.~Nock, and L.~Qu}, {\em Making
  deep neural networks robust to label noise: A loss correction approach}, in
  Proceedings of the IEEE Conference on Computer Vision and Pattern
  Recognition, 2017, pp.~1944--1952.

\bibitem{ren2018learning}
{\sc M.~Ren, W.~Zeng, B.~Yang, and R.~Urtasun}, {\em Learning to reweight
  examples for robust deep learning}, arXiv preprint arXiv:1803.09050,  (2018).

\bibitem{steinhardt2017certified}
{\sc J.~Steinhardt, P.~W.~W. Koh, and P.~S. Liang}, {\em Certified defenses for
  data poisoning attacks}, in Advances in neural information processing
  systems, 2017, pp.~3517--3529.

\bibitem{sukhbaatar2014training}
{\sc S.~Sukhbaatar, J.~Bruna, M.~Paluri, L.~Bourdev, and R.~Fergus}, {\em
  Training convolutional networks with noisy labels}, arXiv preprint
  arXiv:1406.2080,  (2014).

\bibitem{van2015learning}
{\sc B.~Van~Rooyen, A.~Menon, and R.~C. Williamson}, {\em Learning with
  symmetric label noise: The importance of being unhinged}, in Advances in
  Neural Information Processing Systems, 2015, pp.~10--18.

\bibitem{veit2017learning}
{\sc A.~Veit, N.~Alldrin, G.~Chechik, I.~Krasin, A.~Gupta, and S.~Belongie},
  {\em Learning from noisy large-scale datasets with minimal supervision}, in
  Proceedings of the IEEE Conference on Computer Vision and Pattern
  Recognition, 2017, pp.~839--847.

\bibitem{xiao2017fashion}
{\sc H.~Xiao, K.~Rasul, and R.~Vollgraf}, {\em Fashion-mnist: a novel image
  dataset for benchmarking machine learning algorithms}, arXiv preprint
  arXiv:1708.07747,  (2017).

\bibitem{xiao2015learning}
{\sc T.~Xiao, T.~Xia, Y.~Yang, C.~Huang, and X.~Wang}, {\em Learning from
  massive noisy labeled data for image classification}, in Proceedings of the
  IEEE conference on computer vision and pattern recognition, 2015,
  pp.~2691--2699.

\bibitem{zhang2003robustness}
{\sc J.~Zhang and Y.~Yang}, {\em Robustness of regularized linear
  classification methods in text categorization}, in Proceedings of the 26th
  annual international ACM SIGIR conference on Research and development in
  informaion retrieval, ACM, 2003, pp.~190--197.

\bibitem{zhang2018generalized}
{\sc Z.~Zhang and M.~Sabuncu}, {\em Generalized cross entropy loss for training
  deep neural networks with noisy labels}, in Advances in Neural Information
  Processing Systems, 2018, pp.~8792--8802.

\bibitem{zhu2004class}
{\sc X.~Zhu and X.~Wu}, {\em Class noise vs. attribute noise: A quantitative
  study}, Artificial intelligence review, 22 (2004), pp.~177--210.

\end{thebibliography}
